%% file: 0main.tex
\UseRawInputEncoding
\pdfoutput=1
\documentclass[acmlarge]{acmart}
\renewcommand\footnotetextcopyrightpermission[1]{}
\usepackage{bm}
\usepackage{amsmath}
\usepackage{mathtools}
\usepackage{CJKutf8}
\usepackage{caption}
\usepackage{subfigure}
\usepackage{array}
\usepackage{color}
\usepackage{wrapfig}
\usepackage{makecell}

\AtBeginDocument{%
  \providecommand\BibTeX{{%
    \normalfont B\kern-0.5em{\scshape i\kern-0.25em b}\kern-0.8em\TeX}}}

\begin{document}

\title{Cross Vision-RF Gait Re-identification with Low-cost RGB-D Cameras and mmWave Radars}

\author{Dongjiang Cao}
\orcid{0000-0002-0429-1779}
\email{djcao@seu.edu.cn}
\affiliation{
  \institution{Southeast University}
  \city{Nanjing}
  \state{Jiangsu}
  \country{China}
}

\author{Ruofeng Liu}
\orcid{0000-0001-9804-3727}
\email{liux4189@umn.edu}
\affiliation{
  \institution{University of Minnesota}
  \city{Minnesota}
  \state{Minnesota}
  \country{United States}
}

\author{Hao Li}
\orcid{0000-0002-7403-2449}
\email{220194384@seu.edu.cn}
\affiliation{
 \institution{Southeast University}
  \city{Nanjing}
  \state{Jiangsu}
  \country{China}
}

\author{Shuai Wang}
\orcid{0000-0002-3609-2205}
\email{shuaiwang@seu.edu.cn}
\authornote{Corresponding author.}
\affiliation{
  \institution{Southeast University}
  \city{Nanjing}
  \state{Jiangsu}
  \country{China}
}

\author{Wenchao Jiang}
\orcid{0000-0001-6765-2466}
\email{wenchao_jiang@sutd.edu.sg}
\affiliation{
 \institution{Singapore University of Technology and Design}
 \city{Singapore}
 \country{Singapore}
}

\author{Chris Xiaoxuan Lu}
\orcid{0000-0002-3733-4480}
\email{xiaoxuan.lu@ed.ac.uk}
\affiliation{
 \institution{The University of Edinburgh}
 \city{Edinburgh}
 \country{United Kingdom}
}

\renewcommand{\shortauthors}{Cao et al.}

\begin{abstract}

Human identification is a key requirement for many applications in everyday life, such as personalized services, automatic surveillance, continuous authentication, and contact tracing during pandemics, etc. This work studies the problem of cross-modal human re-identification (ReID), in response to the regular human movements across \textcolor{black}{camera-allowed} regions (e.g., streets) and \textcolor{black}{camera-restricted} regions (e.g., offices) deployed with heterogeneous sensors. By leveraging the emerging low-cost RGB-D cameras and mmWave radars, we propose the first-of-its-kind vision-RF system for cross-modal multi-person ReID at the same time. {\color{black} Firstly, to address the fundamental inter-modality discrepancy, we propose a novel signature synthesis algorithm based on the observed specular reflection model of a human body. 
Secondly, an effective cross-modal deep metric learning model is introduced to deal with interference caused by unsynchronized data across radars and cameras.} 
Through extensive experiments in both indoor and outdoor environments, we demonstrate that our proposed system is able to achieve $\sim$ 92.5\% top-1 accuracy and $\sim$ 97.5\% top-5 accuracy out of 56 volunteers. 
We also show that our proposed system is able to robustly reidentify subjects even when multiple subjects are present in the sensors’ field of view.
\end{abstract}



\maketitle
\pagestyle{plain} 

\input{1introduction}

\input{2Motivation}

\input{3Overview}

\input{4Background_and_challenges}

\input{5Opportunity_behind_specular_reflection}

\input{6Similarity_calculation}

\input{7Gait_cycle_extraction_and_denoise}
\input{8Experiment_setup_and_data_collection}

\input{9Performance_Evaluation}
\input{10Related_work}

\input{11Discussion}
\input{12Conclusion}

\bibliographystyle{ACM-Reference-Format}
\bibliography{13references}

\end{document}

%% file: 1introduction.tex
\UseRawInputEncoding
\vspace{-1mm}
\section{Introduction}
\thispagestyle{plain}
 
Person identification is a key enabler to realize ubiquitous computing and finds many applications in everyday life, such as personalized services, automatic surveillance, continuous authentication, and contact tracing during pandemics, etc. A variety of sensing techniques have been proposed for person identification and are used in different scenarios. For example, in \textcolor{black}{camera-allowed} areas where privacy is less a concern, e.g., streets and entrances to buildings, cameras or other vision-based techniques have been extensively adopted for identification. In contrast, radio frequencies (RF) based sensors are widely known to be superior for privacy preservation \cite{zhao2019mid} and are suitable in \textcolor{black}{camera-restricted}  scenarios where any vision-based techniques (including both RGB and depth imaging sensors) will raise severe privacy concerns and legal issues, e.g., home, restroom, hospital, offices, or some confidential areas in a factory.

\textcolor{black}{Camera-allowed} and \textcolor{black}{camera-restricted} regions, on the other hand, are often connected with each other in the real world with people regularly moving across them. Moreover, due to the widespread IoT deployments, both of these regions are being increasingly installed with different kinds of sensors \cite{cao2018enabling,xu2019ivr,liu2018tar}. For instance, it is common for workers in a factory to leave from public spaces (e.g., monitored by RGB or RGB-D cameras) and enter confidential warehouses (e.g, monitored by RF sensors) or vice versa. The movement of humans across scenarios with different privacy considerations and sensor deployments leads to a new need for cross-modal human re-identification (\textbf{ReID}) - \textit{given a person detected by a RF sensor, the system can identify the same person in camera footage or vice versa.} 

Among the wide spectrum of vision-based and RF-based sensors, single-chip millimeter-wave (mmWave) radars and RGB-D cameras respectively emerge as two promising modalities in recent years. Single-chip mmWave radars are a type of sensor designed to have the capability of simultaneously sensing multiple targets in the field of view (FoV), generating precise spatial information (i.e., range and angle of arrival)  of each target \cite{yang2020mu}. Meanwhile, as a RF-based sensing modality, mmWave radars are impervious to any lighting conditions, e.g., darkness, dimness, and glare. These advantages lend the radar itself an effective sensing alternative when cameras fall short \cite{jiang2020mmvib,shuai2021millieye,khamis2020rfwash,xu2021followupar}. On the other hand, RGB-D cameras are a type of depth sensing device that work in association with a RGB camera. As they are able to augment the conventional image with depth and scale information, RGB-D cameras significantly outperform other traditional cameras (e.g., RGB cameras) in the domain of vision-based human sensing \cite{haque2016recurrent}. With the recent maturity of manufacturing, low-cost RGB-D cameras (e.g., Microsoft Kinect and Intel Real Sense) and mmWave radars (e.g., Texas Instruments IWR series) become available which further strengthen their promise for human sensing tasks.

Motivated by the need for cross-modal identification and the emerging opportunity above, this work proposes the first cross-modal re-identification design among RGB-D cameras and mmWave radar. The key idea underpinning our design is leveraging the human gaits (i.e., how a person walks) across modalities, which is a unique and consistent biometric identifier of each individual \cite{nambiar2019gait}. To achieve cross-modal ReID, gait information is captured in both modalities and their similarity is estimated to associate people between cameras and radars.

   
While gait recognition has been studied separately with mmWave radars \cite{yang2020mu} or RGB-D cameras \cite{deng2017fusion,chattopadhyay2014frontal}, new technical challenges come up when leveraging the cross-modal data for human identification. \textit{First}, the sensing data from vision and radar inherently suffer from severe inter-modality discrepancy. \textcolor{black}{Cameras capture detailed shapes and motions of each body part, whereas the radar only provides discrete and sparse detection points, which renders a direct similarity calculation across modalities impossible.} 
\textcolor{black}{
\emph{Second}, as the camera and radar in our context are installed in different places, their data are not  collected at the same time and thus are often not synchronized in fine-granularity. As a result, a reliable similarity estimation must be robust to various interference factors (e.g., temporal misalignment and minor gait changes).
}

In order to address the above technical challenges, several novel designs are proposed in this work. 
\emph{First}, to tackle the inter-modality discrepancy, we propose a modal-unifying method driven by the physical model of specular reflection \cite{adib2015capturing}. We find that the human body acts as a spectral reflector (like a mirror) of incident mmWave signal, and thus only signals arriving close to the normal of the surface on the human body are reflected back towards the receiving antenna \cite{ahmed2012illumination}. Consequently, when a person is moving relative to the radar, the body parts being captured change over time and form spatio-temporal signatures of human gait. Such gait signatures differ significantly from person to person for their different walking styles (e.g., the stride of legs, arms swing and motion of torso). Moreover, the gait captured by the RGB-D cameras and mmWave radars can be tightly corresponded via the specular reflection model which bridges the heterogeneity between vision and RF data. \textcolor{black}{Based on this observation, we design a novel algorithm to synthesize signature points from depth images while addressing imperfections  of raw RGB-D data and view angle difference of the two sensors). \textit{Next}, to deal with interference factors caused by unsynchronized data collections, we design an end-to-end deep metric learning pipeline for similarity estimation. The model extracts high-level representations of gait features from both synthesized signatures and radar point clouds and estimates their similarity in the high-dimensional latent space, thus addressing the temporal misalignment, minor walking speed change, etc. 
}

To summarize, our work makes the following contributions:
\begin{itemize}
    \item We present the first-of-its-kind cross-modal re-identification design between mmWave radars and RGB-D cameras, achieving simultaneous multi-person ReID across \textcolor{black}{camera-allowed and camera-restricted} scenarios. 
    \item  We propose novel designs to address intrinsic technical challenges of cross-modal ReID (e.g., inter-modality discrepancy and \textcolor{black}{similarity estimation}).
    \item We implement our design with commodity mmWave radar (i.e., IWR6843) and RGB camera (i.e., Azure Kinect). Extensive evaluations with 56 volunteers demonstrate that our design is highly accurate (92.5\% top-1 accuracy and 97.5\% top-5 accuracy ) and robust in handling multi-person ReID. This multimodal dataset collection is released to the community.
\end{itemize}

%% file: 2Motivation.tex
\vspace{-1mm}
\section{Motivation}

\subsection{The need for cross Vision-RF identification}

{\color{black}{\bfseries The need for cross Vision-RF identification:} Depending on requirements of specific applications and level of privacy sensitivity in various scenarios, RGB-D cameras and RF sensors are deployed at different scenes.  For example, nowadays public streets in many smart cities are deployed with surveillance RGB-D cameras to achieve fast and accurate behavior recognition and three-dimensional positioning \cite{sun2018multimodal, patwardhan2017hostile}, which overcomes the limitations of traditional RGB camera in dealing with lighting changes, color similarity and shadows.  On the other hand, many people prefer to install RF techniques (e.g., WiFi and radar) to monitor their home \cite{liu2020real,huawei}, which avoids privacy invasion (e.g., room layout) while providing various intelligent services such as smart control \cite{liu2020real} and fall detection for the elderly \cite{sun2019privacy}.  Another example of the hybrid deployment is the enterprise scenario (e.g., office and factory) where RGB-D cameras are increasingly being used in building entrance control systems to improve the  accuracy and robustness \cite{pamplona2013continuous} of identity recognition (e.g., liveness detection \cite{sun2018multimodal}), whereas RF sensors are more acceptable at meeting rooms, offices and laboratories because they prevent the leakage of the enterprise's confidential information. This emerging heterogeneous sensing scenario motivates us to study the possibility of cross vision-RF ReID. Specifically, when a RF sensor detects a person of interest in the camera-restricted area, our design aims at retrieving the same person's footage from cameras installed in camera-allowed areas. We describe three broad sets of applications that this system can add values on:}

\textbf{1) Personalized Services. }
By matching a person detected by RF sensor to his/her image, we could exploit the rich information in the image (e.g., age, gender, and apparel) to provide personalized services without invading their privacy (e.g., information about fine-grained activities or confidential behaviors) in  {\color{black}camera-restricted} areas. For example, we could adjust the temperature in a private office  based on the clothes of  a person captured by cameras at the entrance.

\textbf{2) Tracking and Identification. } 
Our design also enables seamless tracking  when people move between the camera and RF monitored areas, generating complete trajectories across    \textcolor{black}{camera-allowed and camera-restricted} scenes. Seamless tracking across spaces can be realized and assist in situations such as contract tracing during pandemics.

\textbf{3) Security and Surveillance. }
Furthermore, our technique enriches RF sensors' capability on security surveillance.  For example, when trespassing happens at home, RF-based surveillance monitors could not only detect intruders but also utilize wireless signal as the cues to retrieve the image of the intruders from video footage of the camera on a nearby public street. This will help the investigator to quickly identify the suspects.

\vspace{-1mm}
\subsection{Limitation of existing solutions}

\noindent {\color{black} \textbf{Single-modal Person Re-Identification.} Person ReID among homogeneous sensors have been investigated for both vision \cite{wang2020unsupervised} and RF technologies \cite{meng2020gait,pegoraro2020multiperson,wang2020adversarial} \textit{separately}. However, these techniques only provide ReID among the specific regions that are equipped with the same type of sensors. Therefore, they cannot accommodate the ReID problem when the subjectives are moving across camera-allowed and camera-restricted scenarios that are covered by different types of sensors. This fundamental limitation motivates us to design a new technology that can effectively bridge these isolated ReID systems, providing a unified identity solution that serves real-world scenarios that contains areas with different camera restrictions.
}
\noindent {\color{black}  \textbf{Cross-modal Person Identification.} Person ReID across different sensing modality is mainly studied by computer vision community with a focus on different types of cameras, including the identification between depth and RGB images \cite{haque2016recurrent, karianakis2018reinforced}, infrared and RGB images \cite{wu2017rgb,ye2018visible}, and images with different resolutions \cite{li2019recover}. These solutions are only applicable at camera-allowed areas. On the other hand, the cross-modal ReID between vision and RF signals, despite its importance and ubiquity, receives little attention. }
{\color{black}
The recent work (i.e., XModal-ID) \cite{korany2019xmodal} is the pioneering study of cross-modal ReID between the camera and RF. They propose to match video and WiFi signals using channel state information (CSI). Nevertheless, as CSI-based techniques struggle to simultaneously identify multiple people in the same scene, XModal-ID has to assume that there is only one person walking in the WiFi area at a time \cite{korany2019xmodal}. This assumption, however, prohibits it from dealing with multi-person scenarios (e.g., contact tracing). In addition, XModal-ID  requires a separate WiFi transmitter and receiver, which often introduces extra deployment cost and calibration complexity.
}

\subsection{Our proposed approach}
\noindent \textbf{Our proposed approach: } In this work, we utilize the emerging RF and vision sensors (i.e., mmWave radar and RGB-D camera) to achieve ReID between \textcolor{black}{camera-allowed and camera-restricted} areas. Recently, both low-cost commercial mmWave radar and RGB-D camera have become widely available and  a large body of human sensing applications (e.g., people counting, intruder detection, and health monitoring) have been developed with these two sensing modalities \cite{yang2020mu,deng2017fusion}. We envision that both sensors will be massively deployed in the near future. Technically, both mmWave radars and RGB-D cameras provide the precise three-dimension spatial information of multiple targets (e.g., subjects' locations and heights) in the FoV. It is therefore advantageous to exploit this 3D sensing capability to solve multi-person identification problems \footnote{Note that although less fine-grained 3D information can be obtained by a depth imaging sensor alone, we focus on RGB-D cameras in this work because using RGB images to enhance raw depth images has been widely available as an option \cite{depth_camera_tutorial,yeung2021effects,yan2018ddrnet}.}. Finally, because the radar itself is a transceiver device, only one radar is required for human sensing so its deployment is more flexible than WiFi (i.e., requiring separate WiFi transmitter and receiver). The opportunities above motive us to study the cross-modal ReID problem between mmWave radars and RGB-D cameras.

\begin{figure*}[h]
  \includegraphics[width=0.9\linewidth]{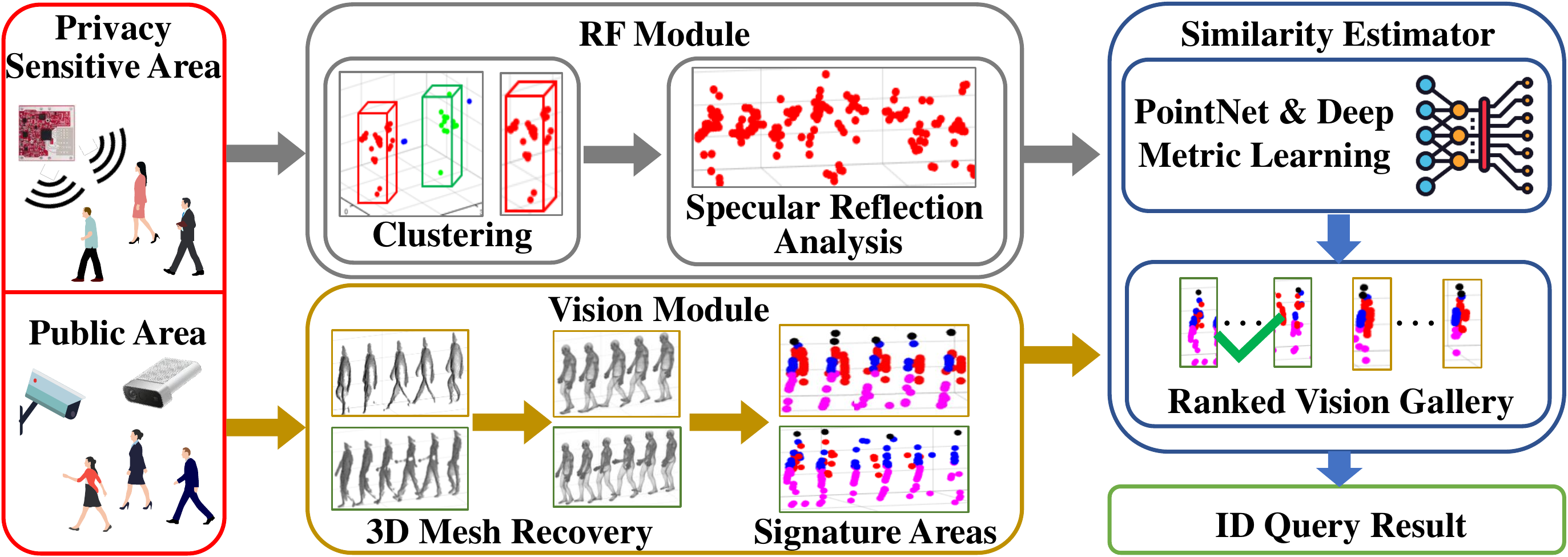}
  \vspace{-2mm}
  \caption{System architecture.}
  \label{fig:System architecture}
  \vspace{-4mm}
\end{figure*}

%% file: 3Overview.tex
\UseRawInputEncoding 
\section{System Overview}

In this paper, we propose a cross vision-RF person ReID system. Given a person detected by a mmWave sensor, the system re-identifies him/her in a gallery of  RGB-D camera footage using the consistent gait features. 
Fig.\ref{fig:System architecture} shows the overall architecture of our design. The RF module processes the received radar signal to obtain the radar points that belong to the person of interest. 
On the other hand, the vision module synthesizes signature areas from the reconstructed 3D mesh of a candidate captured by RGB-D cameras, by using the specular reflection phenomenon of mmWave (Section \ref{sec:opportunity}).  Then, we measure the walking style similarity by comparing the sequence of radar points and synthesized signature areas (Section \ref{sec:design}). The similarity of all the candidates are ranked and the one with the highest similarity score, or those with top similarity scores, are ultimately selected as the output.

%% file: 4Background_and_challenges.tex
\UseRawInputEncoding 
\section{Background}\label{sec:background}
 
This section introduces the primer of mmWave radar, gait, and person ReID needed in this work. We refer the interested readers to \cite{yeung2021effects} for the background of RGB-D cameras and avoid repeating the same content here.
\vspace{-1mm}
\subsection{Principles of mmWave Radar}\label{sec:radarbackground}
 
The single-chip mmWave radar is based on the principles of frequency modulated continuous wave (FMCW) and has the ability to measure the range, relative radial speed and angle of the target.
Specifically, the FMCW radar repeatedly transmits continuous chirp signals for a short period of time which frequency increase linearly with time. When receiving the signal reflected by an object, the radar sensor produces Intermediate Frequency (IF) signal, which is analyzed to obtain the three-dimensional position of the object.

{\bfseries Range Measurement.}
Based on the IF signal, the distance $d$ between the object and the radar is calculated as: \vspace{-1mm}
\begin{equation}
d=\frac{f_{IF}\,c\,T_c} {2\,B}
\end{equation}
where the $c$ is the speed of light, $f_{IF}$ is the frequency of the IF signal, $B$ is the bandwidth swept by chirp, and $T_c$ is the duration of chirp. To measure the range of multiple objects at different ranges, a fast Fourier transform (FFT) is performed on the IF signal (i.e., range-FFT). The result of range-FFT represents the frequency response at different ranges. Thanks to the centimeter level range resolution, it has the ability to detect the position of the torso and limbs.

{\bfseries Angle of Arrival Estimation.}
To depict the exact positions of objects in a spatial Cartesian coordinate system, the angle estimation is indispensable. The mmWave radar utilizes a linear antenna array to estimate the object angle. After emitting chirps with the same initial phase, RF Front-end simultaneously samples from multiple receiver antennas. Because the phases of the received signals are different between receiver antennas, the angle of the reflected signal can be estimated. Formally, the AoA can be calculated as:\vspace{-1mm}
\begin{equation}
\theta = \arcsin \frac{\lambda\,\omega} {2\,\pi\, l}
\end{equation}
where $\omega$ denotes the phase difference, $l$ represents the distance between consecutive antennas and $\lambda$ is the wavelength. Once obtains the range and AoA ($\theta$) of the targets, we can get the exact positions of objects in a spatial Cartesian coordinate system.

\subsection{Person Re-Identification and Human Gait}\label{sec:gaitbackground}

\noindent \textbf{Person ReID: } Originally proposed by the computer vision community, person ReID has been widely studied as a specific person retrieval problem across non-overlapping cameras. Given a query person-of-interest, the goal of ReID is to determine whether this person has appeared in another place at a distinct time captured by a different
camera, or even the same camera at a different time instant. The \emph{query} person can be represented by an image and the large collection of recorded images and/or videos is referred to as a candidate \emph{gallery} \cite{li2018harmonious}. In the rest of this paper, we will follow this general terminology of ReID but focus ourselves on a distinct scenario where a query and the gallery collection are obtained by different sensing modalities (mmWave radar V.S. RGB-D camera) and in different places (\textcolor{black}{camera-allowed areas V.S. camera-restricted areas}). 

\vspace{0.5mm}

\noindent {\color{black}\textbf{Human Gait Identification: }  
As a type of biometric when a person is walking, gait information often carries identity-specific patterns that can be utilized to distinguish human identity \cite{nambiar2019gait}. A gait-based identification system starts from extracting walking features over a data sequence, such as step length, magnitude of arm swing, pace, etc, and then utilizes these features as gait signatures for identity matching. Because walking is a dynamic sequence of movements \cite{nambiar2019gait}, the process of gait feature extraction is often applied over a sequence of gait frames. In this paper, we leverage the sequence frames of walking to extract gait features from two different sensors and utilize the gait features to re-identify the same person across two types of data. }

%% file: 5Opportunity_behind_specular_reflection.tex
\UseRawInputEncoding
\vspace{-1mm}
\section{Feasibility of Cross-modal ReID}\label{sec:opportunity}
\vspace{-0.5mm}
This section investigates the feasibility of cross-modal ReID between RGB-D cameras and mmWave radars. We start by introducing the specular reflection model, and then introduce the important notion of spatio-temporal signature utilized for cross-modal identity matching in subsequent sections.

\begin{figure}[b]\vspace{-6mm}
    \centering
    \subfigure[Specular reflection.]
    {
        \begin{minipage}[t]{0.3\linewidth}
            \centering
            \includegraphics[width=0.6\linewidth]{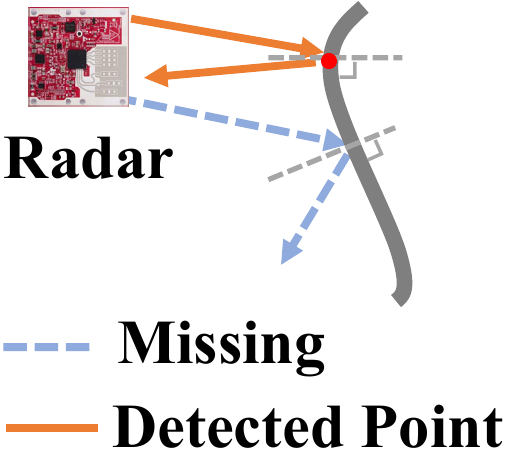}
            \label{}
        \end{minipage}
    }
    \subfigure[Radar detected 
    body areas.]
    {
        \begin{minipage}[t]{0.54\linewidth}
            \centering
            \includegraphics[width=0.6\linewidth]{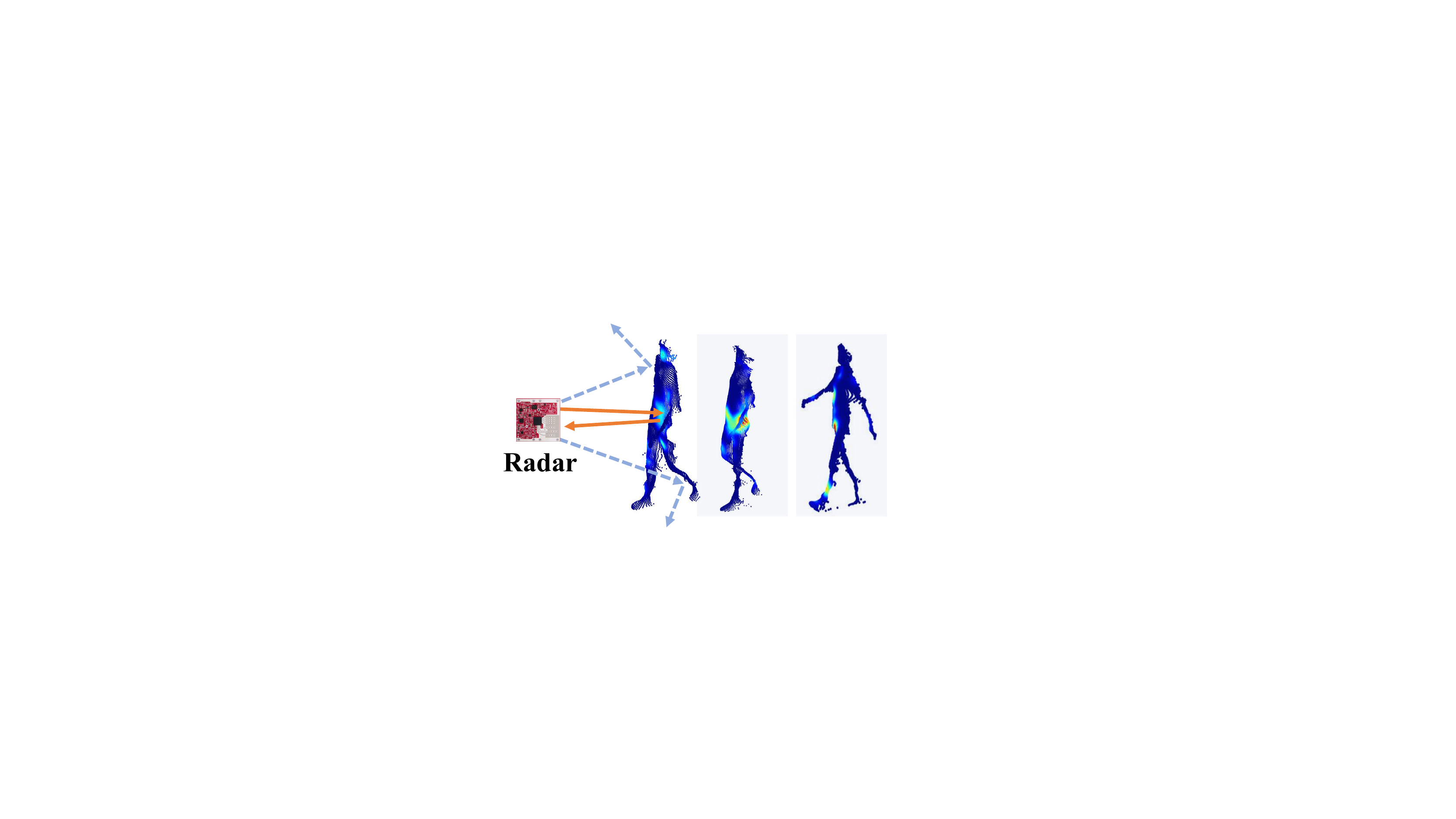}
            \label{}
        \end{minipage}
    }
    \vspace{-3mm}
    \hspace{-3mm}
    \caption{Specular reflection model.}
    \label{fig:specular}
    \vspace{-2mm}
\end{figure}

\begin{figure}[t] \vspace{-3mm}
    \centering
    \subfigure[mmWave radar detection.]
    {
        \begin{minipage}[t]{0.4\linewidth}
            \centering
             \includegraphics[width=0.55\linewidth]{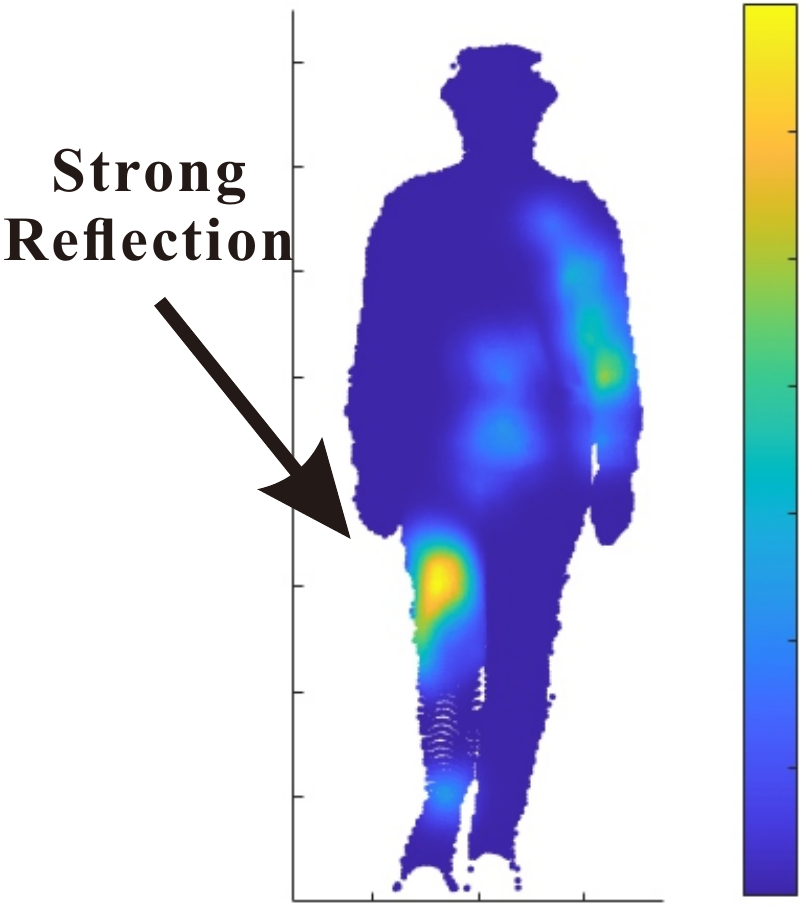}
             \vspace{-4mm}
            \label{}
        \end{minipage}
    }
    \subfigure[Synthesized signature areas.]
    {
        \begin{minipage}[t]{0.4\linewidth}
            \centering
            \includegraphics[width=0.55\linewidth]{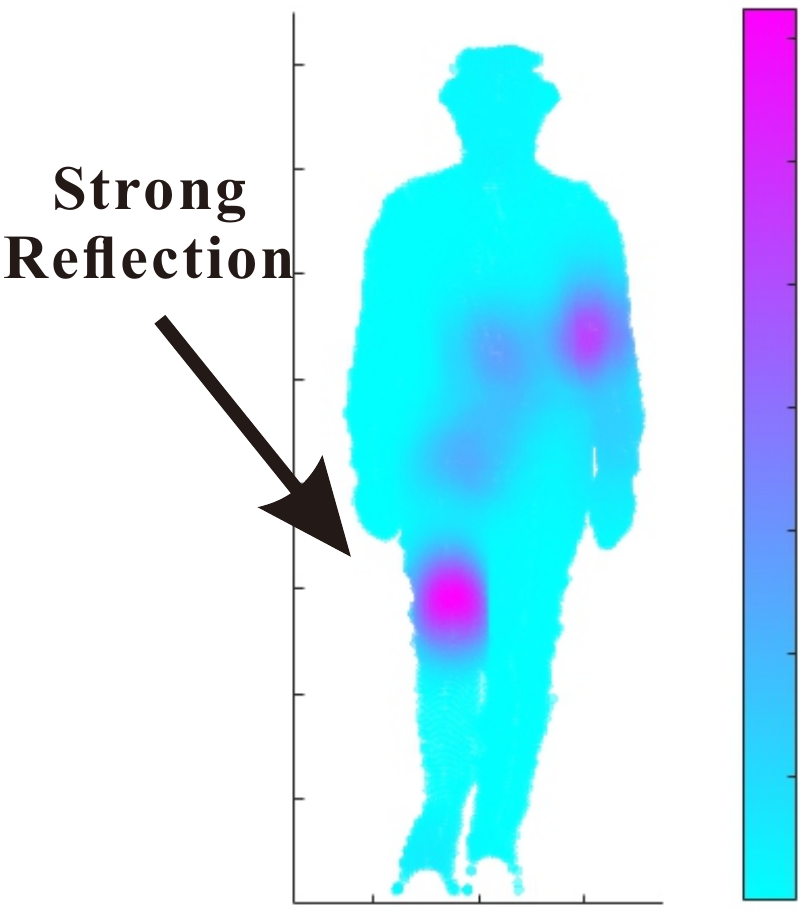}
            \vspace{-4mm}
            \label{}
        \end{minipage}
    }
    \vspace{-4mm}
    \caption{Comparison of radar detection and signature areas on 3D mesh. The intensity of reflection is indicated by brightness. (a) The body areas detected by mmWave radar when a subject is walking. (b) Specular reflection signature areas synthesized based on the specular reflection model. }
    \label{fig-acc-1}
    \vspace{-4mm}
\end{figure}

\subsection{Specular Reflection}\label{sec:specular}

\subsubsection{Specular reflection model}
The critical challenge of cross vision-mmWave ReID lies in the significant data discrepancy between the two modalities. While fine-grained shapes of human bodies can be obtained from RGB-D cameras, a mmWave radar only detects a few points of a target person in a noisy and sparse representation. 


Our key insight to address the discrepancy issue is inspired by the observation that most points detected by a radar are not arbitrary but meaningful reflections on the human body with salient physical characteristics. Specifically, human skin produces strong specular reflections (similar to a mirror) to the incident mmWave signal because the roughness of skin is considerably smaller than the wavelength of mmWave signal and skin also has high water content \cite{ahmed2012illumination}.

As shown in Fig.\ref{fig:specular}(a), when the mmWave signal hits the surface of human skin,  the majority of energy bounces off in the direction symmetric to the incident angle instead of scattering to all directions. Due to the small antenna aperture of a low-cost mmWave radar,  only those signals arriving close to the normal of the surface on the human body can return to the radar, while other signals are deflected away. Consequently, as Fig.\ref{fig:specular}(b) depicts, a radar detects the body area with a unique geometric characteristic - the normal of the surface must point towards the radar device. We hereafter refer to these unique body areas as the ``specular reflection signature areas'' (or ``signature areas'' for short).


\subsubsection{Empirical measurement}
To examine the consistency between signature areas on the human body and the areas detected by a radar, we conduct an empirical measurement. We respectively collect the radar point cloud of 11 subjects from a mmWave sensor as well as their 3D mesh from RGB-D camera (preprocessed with recovery algorithm \cite{zuo2021unsupervised}  as detailed in Section \ref{sec:signature_synthesis}). The signature areas are extracted from the 3D mesh using the specular reflection model (details will be given in Section \ref{sec:signature_synthesis}).   Fig.\ref{fig-acc-1} demonstrates the comparison of the radar detected areas with the signature areas at a radar frame. Strong reflection is indicated by brighter color. Clearly, we can observe that the signature areas reflected from the 3D body mesh largely agree with the points detected by a radar. Both figures in Fig.\ref{fig-acc-1} demonstrate strong reflection on the right thigh (because the right thigh is being lifted) and the left upper arm. 
We further quantify such cross-modal consistency using a basic and intuitive approach - the intersection ratio, i.e., the percentage of radar points that are inside signature areas. Fig.\ref{fig:CCDF_a} depicts the complementary cumulative distribution function (CCDF) of the ratios over 11 people, each of them has 2 records with a total of 50 radar frames (5 seconds). The curves of 11 people show that the majority of radar frames ($\sim$90\%) have over 70\%  radar points within the signature areas. This result motivates us to utilize the specular reflection model to associate the vision and RF data.

\begin{figure*}[t] \vspace{2mm}
  \centering
  \includegraphics[width=0.8\textwidth]{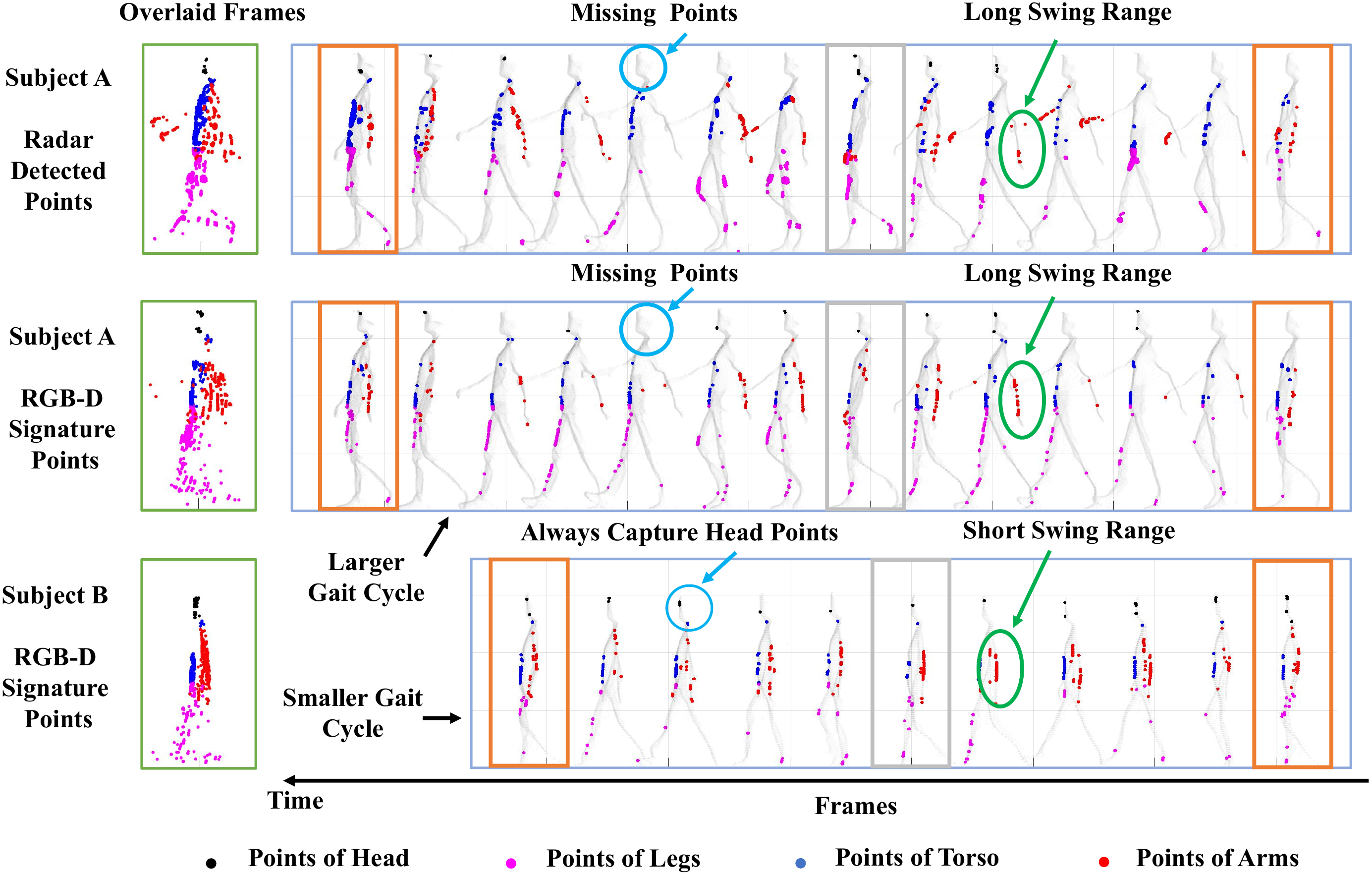}
  \vspace{-2mm}
  \caption{Radar points and spatial-temporal signatures of two subjects. Radar detected points of subject A ($1^{st}$ row). Signature points on 3D mesh of subject A ($2^{nd}$ row). Signature points on 3D mesh of subject B. ($3^{rd}$ row).}
  
  \Description{insight2.}\vspace{-1mm}
  \label{fig:signatureDiff}
  \vspace{-2mm}
\end{figure*}
\subsection{Spatio-temporal Signature of Gait}\label{sec:signature}
\subsubsection{Spatio-temporal Signature}

We exploit the specular reflection model to compare the walking pattern of different people. In specific, when a human is walking towards the radar, the orientation of each body part w.r.t the radar (the angle formed by each part of the body and the radar) changes over time. Therefore, the signature areas (i.e. the subset of body parts detected by the radar) also temporally alternate. 
Notably, as each person has distinct body shapes (e.g., height and body part lengths) and unique gait (e.g., gait period, stride length, and the angle of arm swing), walking towards a radar will produce a sequence of signature areas that possesses a spatio-temporal pattern unique to human identity. It is this unique spatio-temporal pattern that allows us to locate the consistent person's ID across vision and RF modality. We hereafter refer to this consistent ID as ``spatio-temporal signature''.
Fig.\ref{fig:signatureDiff} demonstrates the side views of spatio-temporal signatures of two different subjects during a gait cycle. For ease of visualization, each signature area is aggregated and depicted as a point and the signature areas on different body parts are plotted with different colors. The first and second row of the figure depicts the radar points and signature points of subject A, while the last row is the signature points of subject B. To make it more intuitive, the true mesh of a body obtained from the RGB-D cameras is also given in the grey background.

\vspace{1mm}
\noindent \textbf{Observation 1: } Several notable gait discrepancies among two subjects can be observed from two signatures, including dynamic kinematic characteristics and static shape characteristics e.g.,  1) \emph{stride:} The stride or step length could be observed by trajectories of the leg (pink points). Both the radar and signature points of subject A show a larger step length than subject B. 2) \emph{arm swing:} The distribution of the red points represents the amplitude of arm wings. The swing amplitude of subject A is visibly larger than subject B, as seen from the overlaid frames. 3) \emph{gait cycle:}  
Fig.\ref{fig:signatureDiff} shows a sequence of frames on the whole gait cycle. Looking into the gait cycle, Subject A has a longer gait cycle than Subject B, implying that Subject B has a faster cadence. 4) \emph{height:} The heights of both subjects can be obtained from the range of signature areas in the vertical direction. Also, the head of subject B is detected in all radar frames due to the person's unique height while subject A's head is detected intermittently.

\begin{figure}[b]\vspace{-3mm}
    \centering
     \subfigure[  CCDF of the intersection ratios of 11 people, legend omitted for readability.]
    {
        \begin{minipage}[t]{0.45\linewidth}
            \centering
            \includegraphics[width=0.8\linewidth]{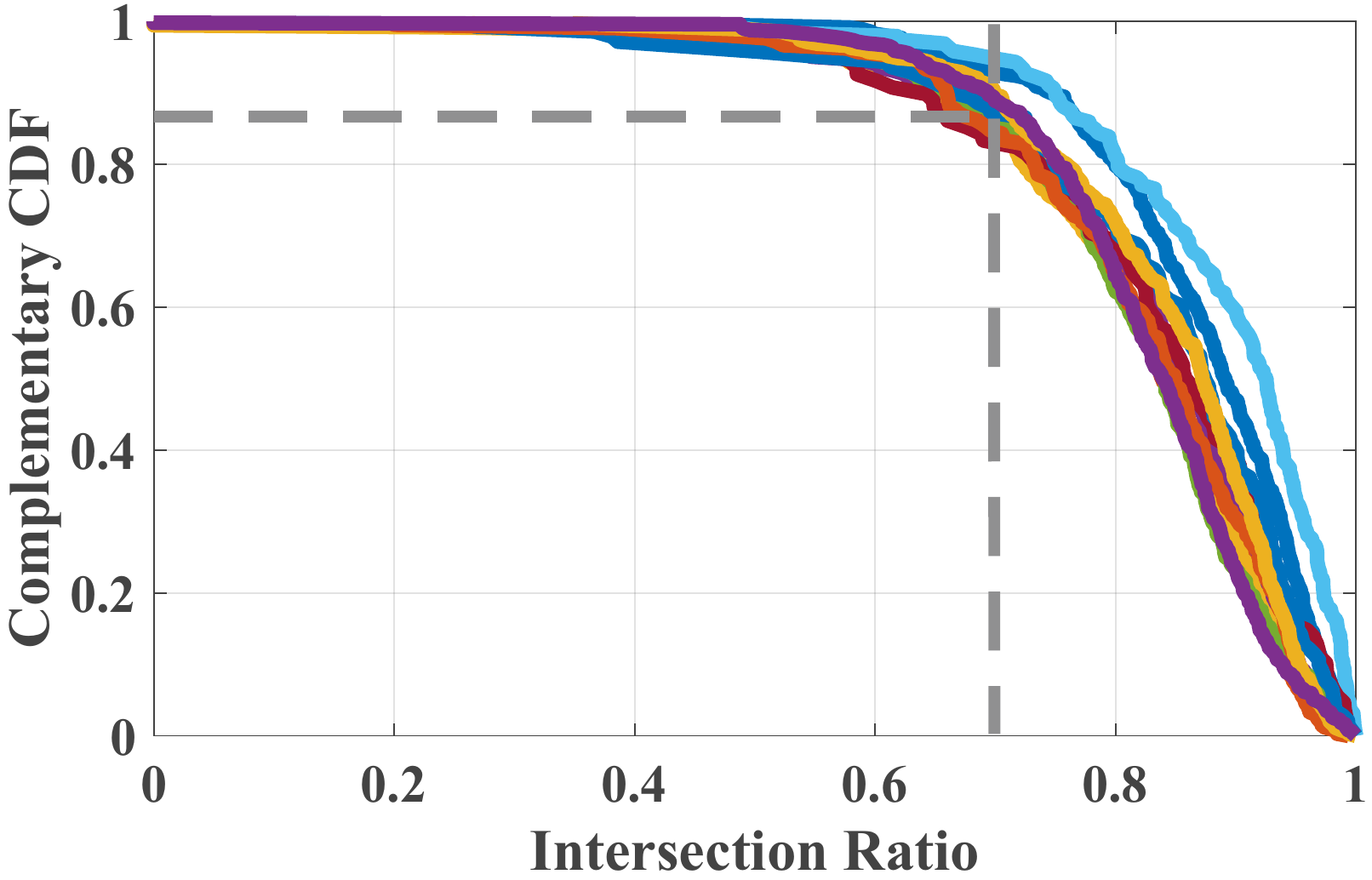}
            \label{fig:CCDF_a}
        \end{minipage}
    }
    \subfigure[ CCDF of the intersection ratios: same person vs. different person]
    {
        \begin{minipage}[t]{0.45\linewidth}
            \centering
            \includegraphics[width=0.8\linewidth]{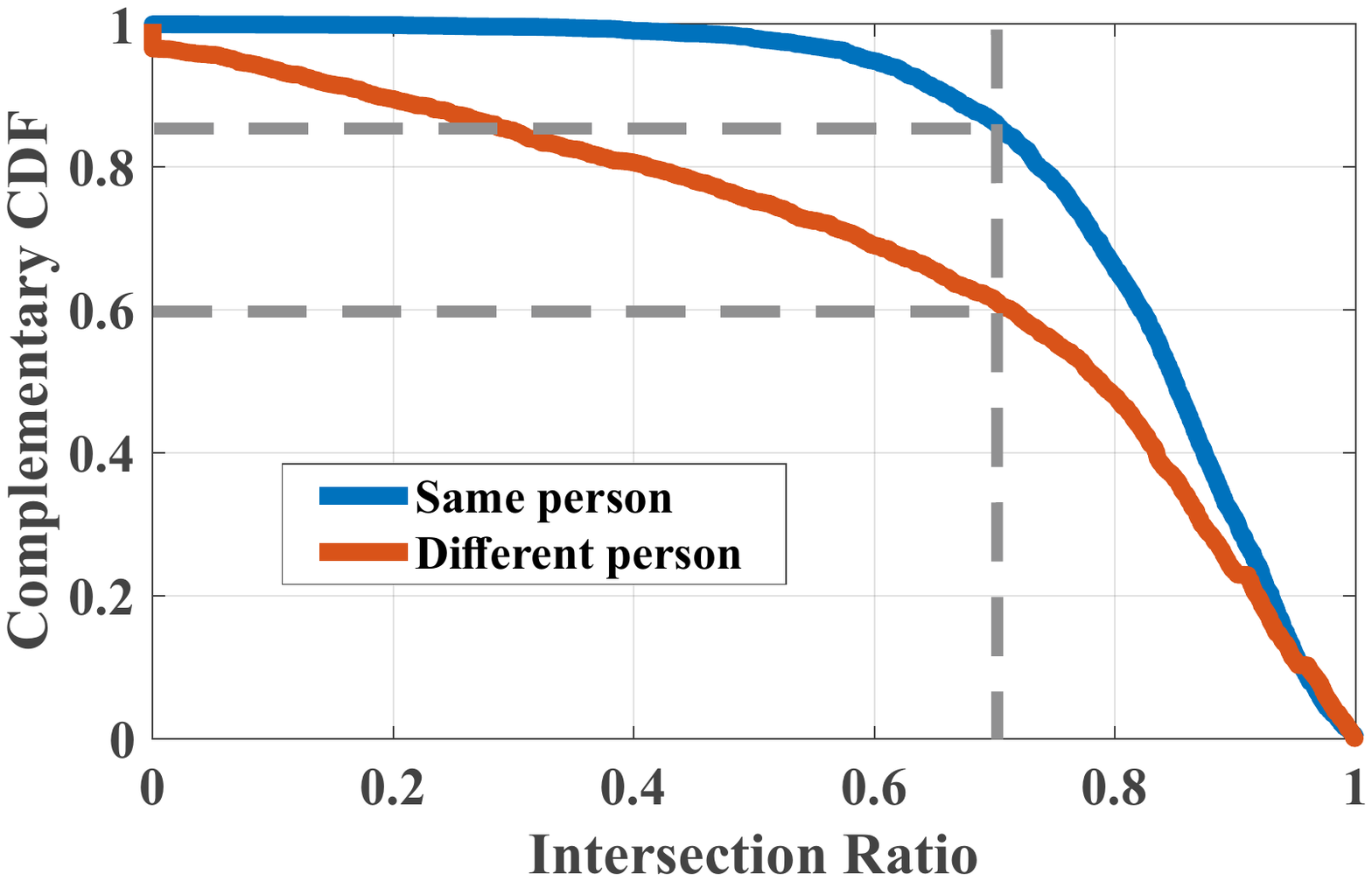}  
            %
            \label{fig:CCDF_Reflection}
        \end{minipage}
    }
    \vspace{-5mm}
    \caption{CCDF of the intersection ratios. (a) Ratios of a same person  (b)  Same person (mean = 82.53\%) vs. different persons (mean = 67.39\%).}
    \label{sec:CCDFsignature}
    \vspace{-4.5mm}
\end{figure}

\vspace{1mm}
\noindent \textbf{Observation 2: } It is also observed that signature areas on various body parts have uneven contributions in differentiating subjects. As Fig.\ref{fig:signatureDiff} shows, for both subjects, the torso can be detected by the radar at a similar horizontal position in almost every frame. This, however, means that the signature areas on the torso carry little identity-specific information.  In contrast, the patterns of signature areas on the arm and leg vary dramatically from person to person because the limbs have a larger degree of changes than the torso when a person is walking and thus can capture the most salient gait features (e.g., step length and arm swing). 


\subsubsection{Empirical measurement}\label{sec:diffbody}

We now quantitatively examine the feasibility of using the signature areas to differentiate different people across vision and radar. Specifically, we analyze the intersection ratio (the percentage of radar points that are inside signature areas) of 11 people with each of them having 50 frames ($\sim4$ gait cycles).  Fig.\ref{fig:CCDF_Reflection} shows the result. The blue curve depicts the CCDF of intersection ratios when the radar points and signature areas are from the same person. The orange curve is the average result of different people. As shown in Fig.\ref{fig:CCDF_Reflection}, for the case of different people, only 60\% of radar frames have over 70\% intersection ratios with the signature areas - a level significantly lower than the intersection ratio measured on the same person ($>$ 86\%). 
\begin{wrapfigure}{r}{4cm}
\vspace{-3.5mm}
\centering
\includegraphics[width=0.2\textwidth]{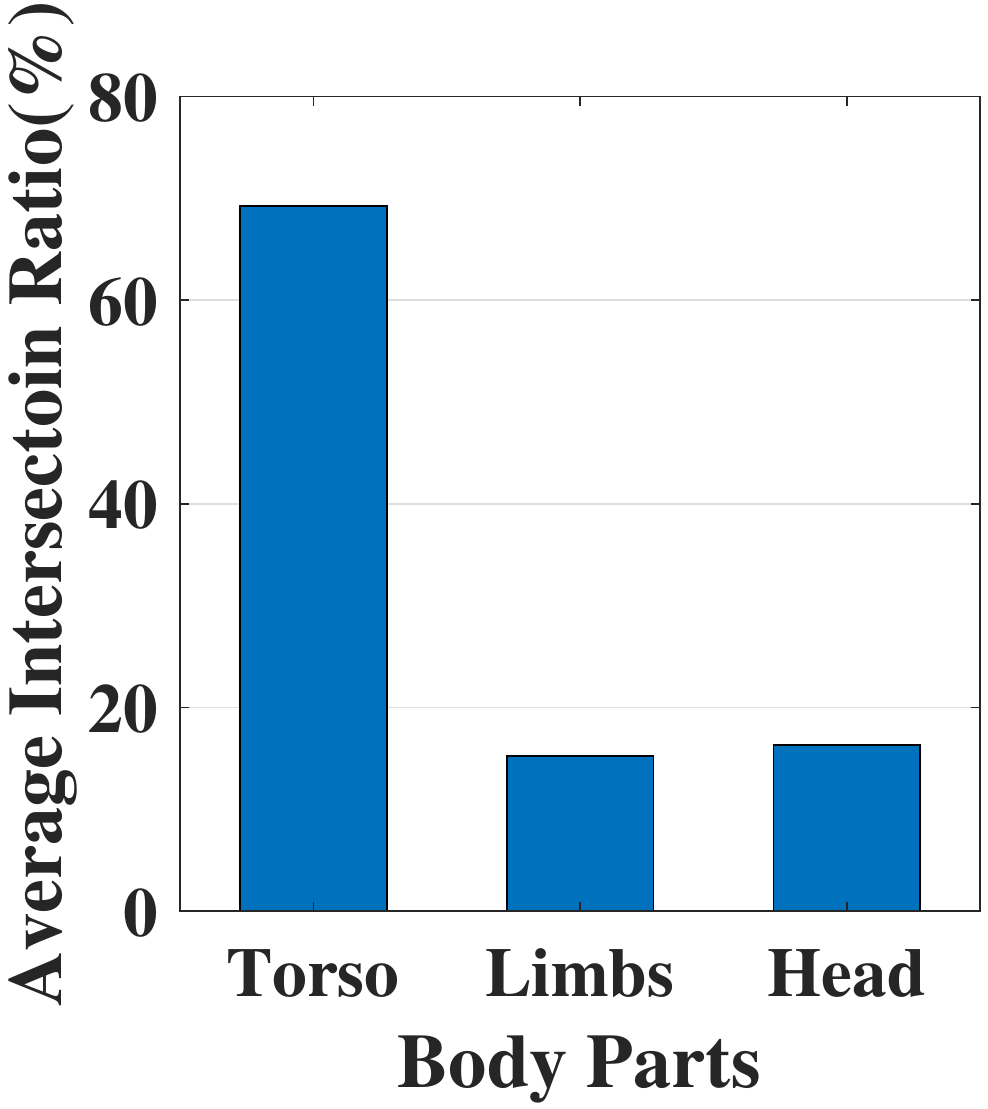}
\vspace{-3mm}
\caption{\footnotesize Comparison of average intersection ratios of different body parts from different people.}
\vspace{-2mm}
\label{fig:fig7}
\Description{}
\end{wrapfigure}  In addition, there are $\sim$25\% radar frames that have less than 50\% intersection ratio between different people. In contrast, this number reduces from 25\% to $<$5\% for the case of the same person,
demonstrating that the effectiveness of the spatio-temporal signature for cross vision-RF ReID. 
Fig.\ref{fig:fig7} further shows the average intersection ratio for these 11 different persons, breakdown in different body parts. As we can see, the intersection ratio of the torso across different people is significantly larger than the other two, implying that the torso should not be considered as salient area when differentiating persons. We hypothesize that it is the torso that contributes the most false intersection ratios among different people in Fig.\ref{fig:CCDF_Reflection}.
Limbs, on the other hand, have a much smaller intersection ratio across different people, indicating their potential effectiveness and should have more weights compared with the torso.

\subsection{Summary and Challenges}

In summary, cross-modal ReID is feasible based on the above observations and empirical studies: 1) specular reflection can accurately model radar detection of human body; 2) a unique spatio-temporal signature can be produced from 3D mesh using the observed specular reflection model, which captures distinct gait features of the subject. Also, signature areas on various body parts have uneven contributions in differentiating subjects; 3) specular reflection can effectively bridge radar and vision modality for cross-modal ReID.

\color{black}
However, to develop a practical ReID algorithm based on these observations, we have to further address several technical challenges. First, we assume a perfect 3D mesh of a subject in spatio-temporal signature synthesis, whereas the raw outputs of a RGB-D camera often suffer from self-occlusion and noise (Section \ref{sec:signature_synthesis}). Second, the 3D mesh is obtained from the camera's view angle. To synthesize the radar signature, we need to transform it to the radar coordinate system, which requires the knowledge of the location and orientation of the subject (Section \ref{sec:signature_synthesis}) in radar. Third, RGB-D image and radar point clouds are collected in a unsynchronized manner. The similarity estimation algorithm needs to be robust against  the temporal misalignment and minor gait difference when the subject walks in disjointed areas (Section \ref{sec:similarity}). Finally, since various body parts contain different amounts of identify-specific information, the algorithm is expected to dynamically adjust their contribution to the similarity estimation  (Section \ref{sec:similarity}).

\color{black}


%% file: 6Similarity_calculation.tex
\UseRawInputEncoding 
\section{Cross Vision-RF Gait Re-identification} \label{sec:design}
This section introduces the design of cross-modal ReID. We start by introducing the signature synthesis based on specular reflection model, and then discuss the similarity estimation.

\subsection{Signature Synthesis}\label{sec:signature_synthesis}
 
{\color{black}\textbf{Mesh Preprocessing.}} The feasibility analysis in Section \ref{sec:specular} indicates that the radar capture has a unique geometric characteristic due to the specular reflection property of human skin. Based on this phenomenon, one can figure out these signature areas on the 3D mesh of a human body and thereby synthesize spatio-temporal signatures for cross-modal ReID. However, using the body mesh obtained from a single RGB-D camera to synthesize the signature has two issues: (i) The RGB-D camera only outputs the surface of the human body directly facing the device while other surfaces are self-occluded. Since the camera and mmWave radar are installed in different locations, they may view the person from different angles. Therefore, synthesizing the radar signature requires those self-occluded body mesh, which is missing in the raw RGB-D output. (ii) The raw output from RGB-D camera contains a lot of holes and noisy points due to undesired artifacts \cite{yan2018ddrnet}. Without handling, these noises and holes could lead to errors in the spectral reflection synthesis. 

\begin{figure}[h]\vspace{-3mm}
    \hspace{-2mm}
    \makeatletter\def\@captype{figure}\makeatother
    \begin{minipage}[h]{0.54\linewidth}\vspace{3.5mm}
            \centering
            \includegraphics[width=1\linewidth]{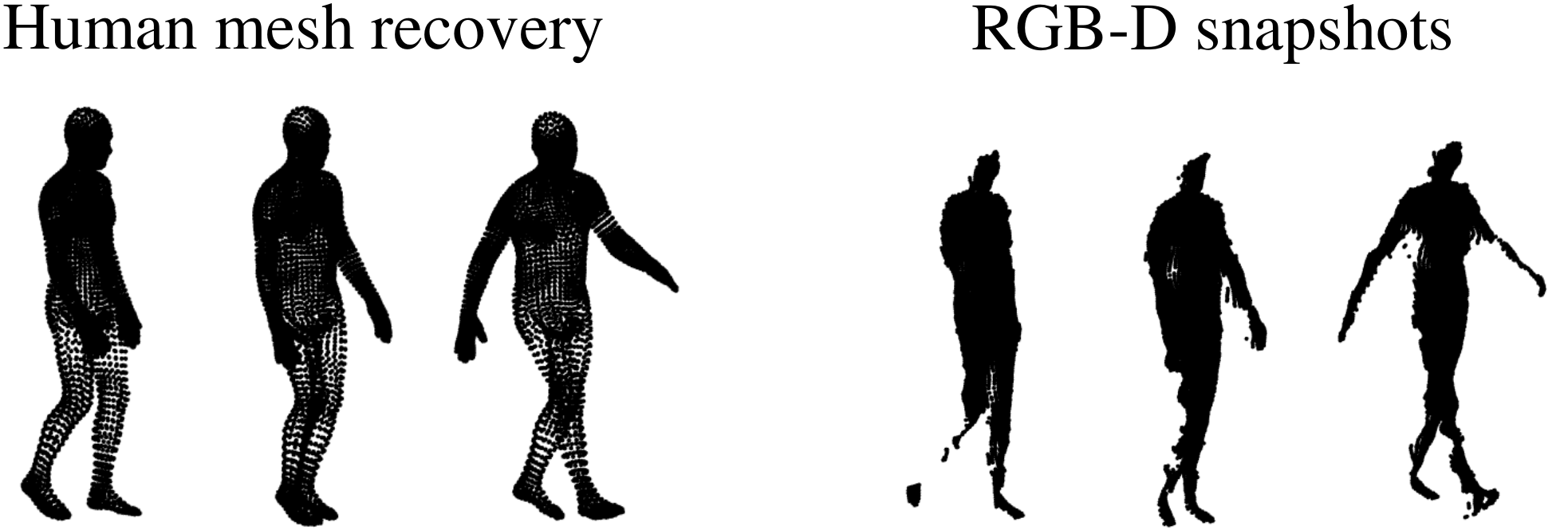}
            \vspace{-6mm}
            \caption{Samples of meshes output from recovery algorithm (left) for snapshots of a walking person (right).}
            \label{fig:recon}
    \end{minipage}
    \hspace{2mm}
    \makeatletter\def\@captype{figure}\makeatother
    \begin{minipage}[h]{0.4\linewidth}
        \centering
        \includegraphics[width=0.8\linewidth]{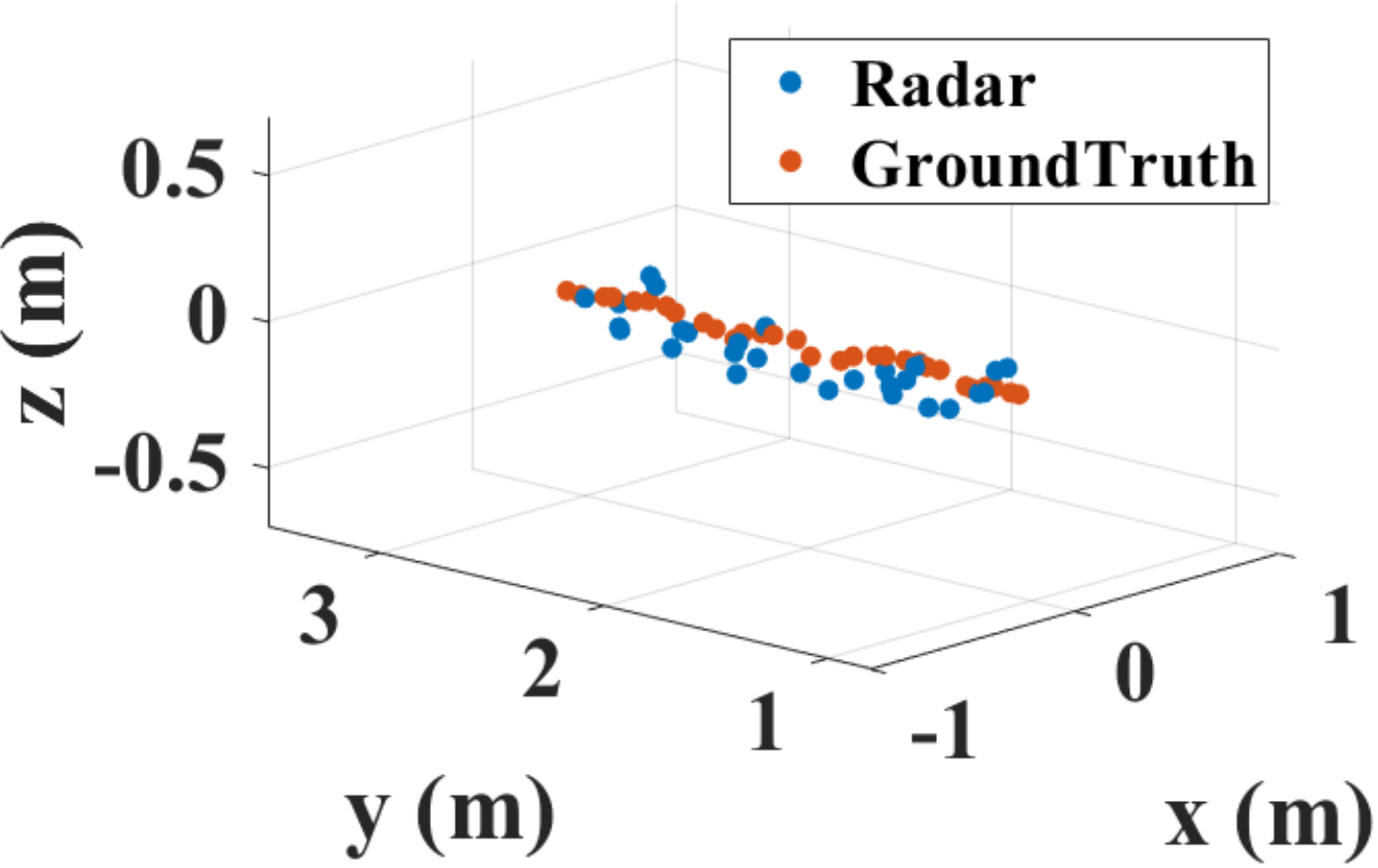}
        \vspace{-2.5mm}
        \caption{Estimation of location and trajectory of subject's center using radar vs. GT using RGB-D.}
        
        \label{fig:heart}
    \end{minipage}
    \vspace{-0.5mm}
\end{figure}

To address these issues, we propose to reconstruct the full-body mesh first from raw RGB-D output before any cross-modal matching is performed. Concretely, we input each frame of RGB-D capture with self-occlusion and noise into the state-of-the-art human mesh recovery algorithm \cite{zuo2021unsupervised}. The algorithm fits the data into a human body model that recovers the self-occluded body mesh and gets rid of the noise. Fig.\ref{fig:recon} shows a few raw RGB-D camera snapshots and their corresponding reconstructed full-body 3D mesh, demonstrating that the algorithm can effectively recover the missing body mesh and reduce the noise.


{\color{black} 
\noindent\textbf{Coordinate Transformation.}
Before using the reconstructed full-body 3D meshes to synthesize the signature areas,  we must first shift the 3D mesh to the cartesian coordinate system of radar. To achieve this, we exploit the rich 3D information of radar to find the location and orientation of the subject. Specifically, we first obtain a rough estimation of the initial position of the subject directly from the first frame of radar point clouds. Then, we fine-tune the position by matching the geometric center of 3D mesh to the one of the radar point cloud, which are figured out by stacking the radar frames in the walking direction.  Furthermore, we estimate the orientation of the subject from a short walking trajectory, which could be considered as a straight line for a very short time. With the initial location and trajectory, we can shift the sequence of 3D mesh to the radar coordinate by mapping the first mesh to the location of the first radar frame and aligning its orientation. Other meshes are also shifted using the first one as the reference. As Fig. \ref{fig:heart} demonstrates, the estimation of location and trajectory is very close to the ground truth. It's worth noting that although our estimation is coarse-grained, small error can be tolerated by the redundancy of signature synthesizer and resiliency in our deep neural network based similarity estimator.
}

\vspace{2mm}
{\color{black}\noindent \textbf{Signature Point Synthesis.} We utilize spectral reflection model to synthesize the radar signature. Note that RGB-D cameras are digital sensors, so our algorithm obtains the discrete samples of these signature areas, which are referred to as ``signature points'' in the following sections. Formally, we denote the set of shifted dense 3D mesh points of a human body at time $t$ by $\mathcal{P}(t)$, while the location of the (virtual) mmWave radar is denoted by $\bm{x}_m = (0,0,0) \in{{\mathcal{R}^3}}$. For each point in $\mathcal{P}(t)$, we find two vectors - the normal vector of the point and the vector from the point to the radar. The normal vector  {$\bm{n}_s(t)$} of the point $\bm{s}(t)$ is determined by the local plane fitted of six neighboring points. And the vector from the point to the radar is $(\bm{s}(t)- \bm{x}_m)$. If the angle between the two vectors is sufficiently small, then that point is considered as a signature point. As shown in equation \ref{eq:specular}, we obtain the spatio-temporal signature $\mathcal{P}_s(t)$, which is a set of signature points. $\epsilon$ is the threshold of the angle, which is set to a non-zero value to tolerate small errors in the subject localization.}

\begin{equation}\hspace{14mm}\vspace{-5mm}
  \mathcal{P}_s(t) = \lbrace \bm{s}(t)\in{\mathcal{P}(t) |  \arccos \frac{(\bm{s}(t)- \bm{x}_m) \cdot \bm{n}_s(t)} {|\bm{s}(t) - \bm{x}_m| |\bm{n}_s(t)|} < \epsilon } \rbrace
  \label{eq:specular}
\end{equation}\vspace{-2mm}

\subsection{Similarity Estimation}\label{sec:similarity}

{\color{black}
\subsubsection{Similarity Estimation via Deep Metric Learning.}  Section \ref{sec:signature} shows that we can synthesize mmWave signature points from a subject's RGB-D images which demonstrates significantly higher similarity with the subject's own radar point cloud  than the one synthesized using the images from a different person. We leverage the key insight to reidentify the radar target among the RGB-D candidates. This section aims at design a reliable estimation algorithm that satisfies the following objectives. First, the radar and RGB-D data are collected in the disjointed areas and thus are not synchronized in the time. Because an explicit alignment of two modalities is non-trivial due to the sparsity of point clouds, we desire that the estimation algorithm can be invariant to the unknown temporal misalignment. Second, the walking speed and step length might slightly change when people are walking in different areas. Our algorithm must be robust against these minor gait variations. Finally, as we introduce in Section \ref{sec:diffbody}, various body parts (e.g., torso, head and limbs) carry different amounts of identity-specific features. Presumably, an effective similarity estimation needs to be anatomy-weighted -  body parts containing more salient gait features (e.g., head and limbs) should have more impact on the similarity estimation result than the parts with less salience (e.g., torso). 

To achieve these goals, we design a deep metric network to estimate the similarity, which have several benefits. The neural network can extract spatio-temporal features from a sequence of point cloud as well as synthesized signature points and map them into the same feature space, which addresses unknown time misalignment. In addition, by comparing the extracted feature instead of the original input data (e.g., by the location of point cloud), the estimator is more tolerant of minor changes in gait and imperfection in the signature synthesis. Finally, the neural network can learn to dynamically adjust the weight of various body parts and frames based on the context (e.g., pose of the subject). 

\begin{figure*}[h] \vspace{-2mm}
    \centering
    \includegraphics[width=0.92\linewidth]{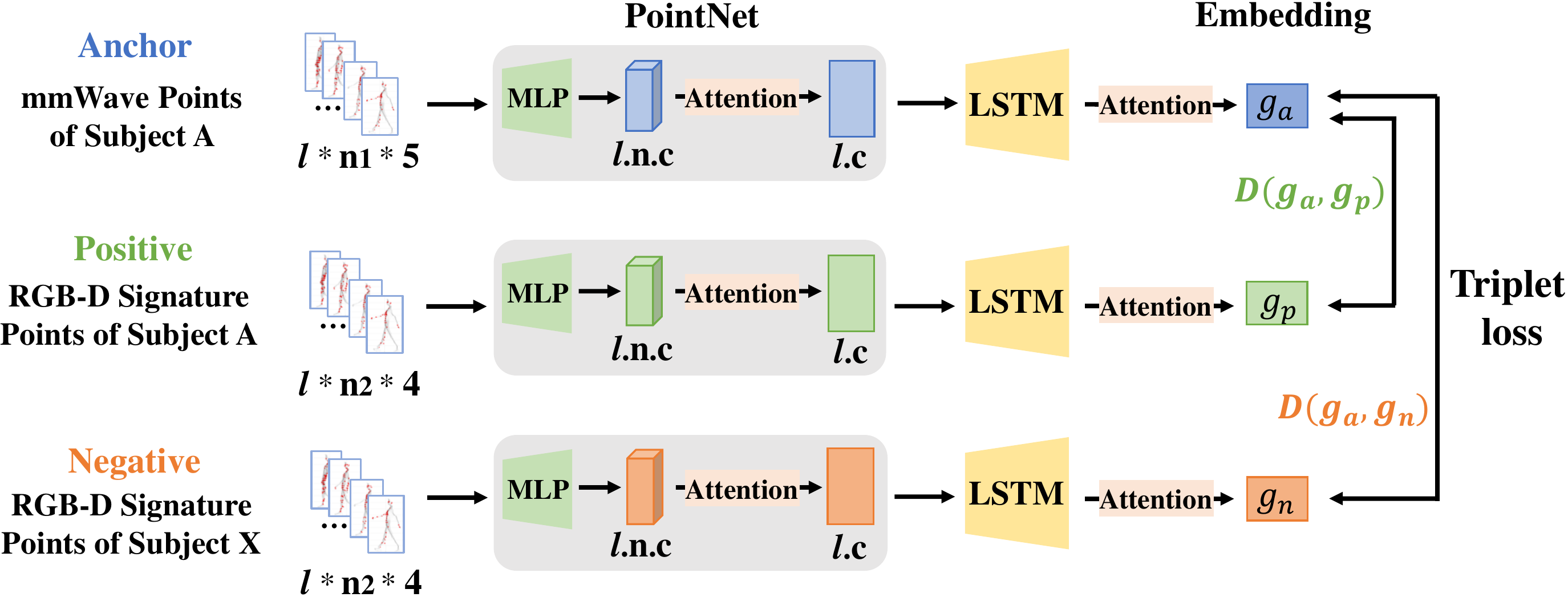}
    \vspace{-1mm}
    \caption{Similarity Estimation via Deep Metric Learning.}
    \label{fig:similarity_estimation}
    \vspace{-4mm}
\end{figure*}

\subsubsection{Similarity Estimation Model}

Fig. \ref{fig:similarity_estimation} depicts the architecture of our designed similarity estimation network based on deep metric learning. Deep metric learning \cite{ge2018deep} is to learning deep feature embeddings that better fits a simple distance function such as Euclidean distance or cosine distance, which has been an active research topic in computer vision community. Given a deep metric model, samples
with similar content are projected onto neighboring locations on a manifold, while samples with different semantic context are mapped apart from each other. As for this work, We train the network with triples:  an anchor instance (i.e., mmWave points of a subject), a positive instance (i.e., RGB-D signature points of the same subject), and a negative instance (i.e.,  RGB-D signature points from a different subject). The network learns to map the both mmWave and RGB-D signature into an identical feature space in which the distance between radar point cloud and signature points of the same person is minimized whereas the distance is maximized when they come from different subjects.

\vspace{2mm}
\noindent\textbf{Feature Extraction: mmWave.}  For mmWave point cloud, we first extract the feature from each individual radar frame with PointNet \cite{qi2017pointnet}. PointNet directly ingests point cloud   $P_t = \lbrace p_{i,t} \rbrace _{i=1} ^N $ contains $N$ mmWave points each of which consists of five features, i.e., 3D coordinates $(x_{i,t},y_{i,t},z_{i,t})$, intensity $(s_{i,t})$ and velocity $(v_{i,t})$. Then, it utilizes permutation-invariant operators (e.g., pointwise MLP and attention) to deal with these unordered points.  More specifically, it encodes each point $p_{i,t}$ independently using a shared-weighted MLP (Multi-layer Perception) and outputs a high-level representation of each point denoted as $m_{i,t} = MLP(p_{i,t}; \theta_m)$, where $\theta_m$ is the learnable parameters of the MLP. Then, we aggregate the features of all the points in a frame using attention mechanism \cite{vaswani2017attention}. The attention computes a score for each point and then calculates a weighted sum of all scores.   Benefited from this mechanism, the estimator  dynamically adjusts the contribution of each point based on the amount of identity-specific features the corresponding body part carries. The attention procedure is formally given as follows: \vspace{-2mm}
\begin{equation}
f_t= \sum_{i= 1}^{N} A_p(m_{i,t};\theta_a) \times m_{i,t}
\end{equation} where  $N$ is the number of points in $t^{th}$ frame, $A_p()$ is a learnable attention function implemented by a full connection layer, and $\theta_a$ is the parameter of attention function. 

Since the gait is a sequence of motion, we need to further exploit the spatio-temporal correlation among each individual frames. To achieve this, we pass the sequence of the extract feature $f_t$ throughput  a LSTM (Long short-term memory network) and  obtain $r_t=LSTM(f_t,r_{t-1};\theta_r)$, where $\theta_r$ denotes the parameters of LSTM. Finally, in order for the result to be invariant to unknown temporal misalignment between two inputs, we aggregate the LSTM output of the all frames using another attention network. As equation \ref{eq:attention2} illustrates, the attention function  computes a
score for each frame in a walking sequence and then aggregates them by a weighted sum of the all scores where the weights are dynamically adjusted based on the significance to the identification.\vspace{-1mm}
\begin{equation}
g_a= \sum_{i= 1}^{L} A_g(r_{t};\theta_g) \times r_{t} \label{eq:attention2}
\end{equation} where $L$ is the length of frame sequence, $\theta_g$ is the parameter of attention function and $g_a$ is the gait feature embedding of mmWave points of an anchor instance.

\vspace{2mm}
\noindent\textbf{Feature Extraction: RGB-D.} We adopt a similar feature extractor for RGB-D signature points. In addition to the pointNet and attention network designed for mmWave, we further exploit the availability of rich amount of information from RGB-D camera to accurately identify the body segment label of each signature point. By doing this, we can guide the network to learn the different salience of body parts in identification task and adjust the weights to the final results.
Therefore, the input RGB-D signature point $q_{i,t}$ is a vector of four features, i.e., 3D coordinates $(x_{i,t},y_{i,t},z_{i,t})$, and body parts index $(b_{i,t})$. We obtain the gait feature embedding of positive and negative RGB-D signature points denoted as $g_p$ and $g_n$.

\vspace{2mm}
\noindent\textbf{Loss Function.} \label{sec:triplet}
Each input of the deep metric objective function is a triple consist of anchor, positive, negative instance denoted as $<g_a, g_p, g_n>$. In order for the samples with the same identities to be as close as possible in the embedding space, and the samples with different identities to be as far away as possible, the objective function is designed as follows: 
\begin{equation}
 L = max(D(g_a,g_p)+margin-D(g_a,g_n),0)
 \label{equ:loss}
\end{equation}
where $D(g_a,g_p)$ is the euclidean distance between embedding of anchor(e.g., mmWave points of the subject) and embedding of positive (e.g., RGB-D signature points of the same subject), $D(g_a,g_n)$ is the distance between embedding of anchor and embedding of negative (e.g., RGB-D signature points of subject X). $margin$ is the hyper-parameter, which makes the distance value of anchor and negative samples larger, while making the distance value of anchor and positive samples smaller. The whole framework is trained end-to-end.

}

%% file: 7Gait_cycle_extraction_and_denoise.tex
\UseRawInputEncoding

%% file: 8Experiment_setup_and_data_collection.tex
\UseRawInputEncoding
\section{Implementation and data collection} \label{sec:implementation}

This section presents the implementation to validate our proposed methodology. We first introduce the experimental setup for collecting data (both of radar and RGB-D camera) and then introduce the preprocessing method of radar data.
\subsection{Experiment Subjects}
\begin{figure}[htbp]\vspace{-5mm}
    \centering
    \subfigure[Age groups.]
    {
        \begin{minipage}[t]{0.48\linewidth}
            \centering
            \includegraphics[width=0.8\linewidth]{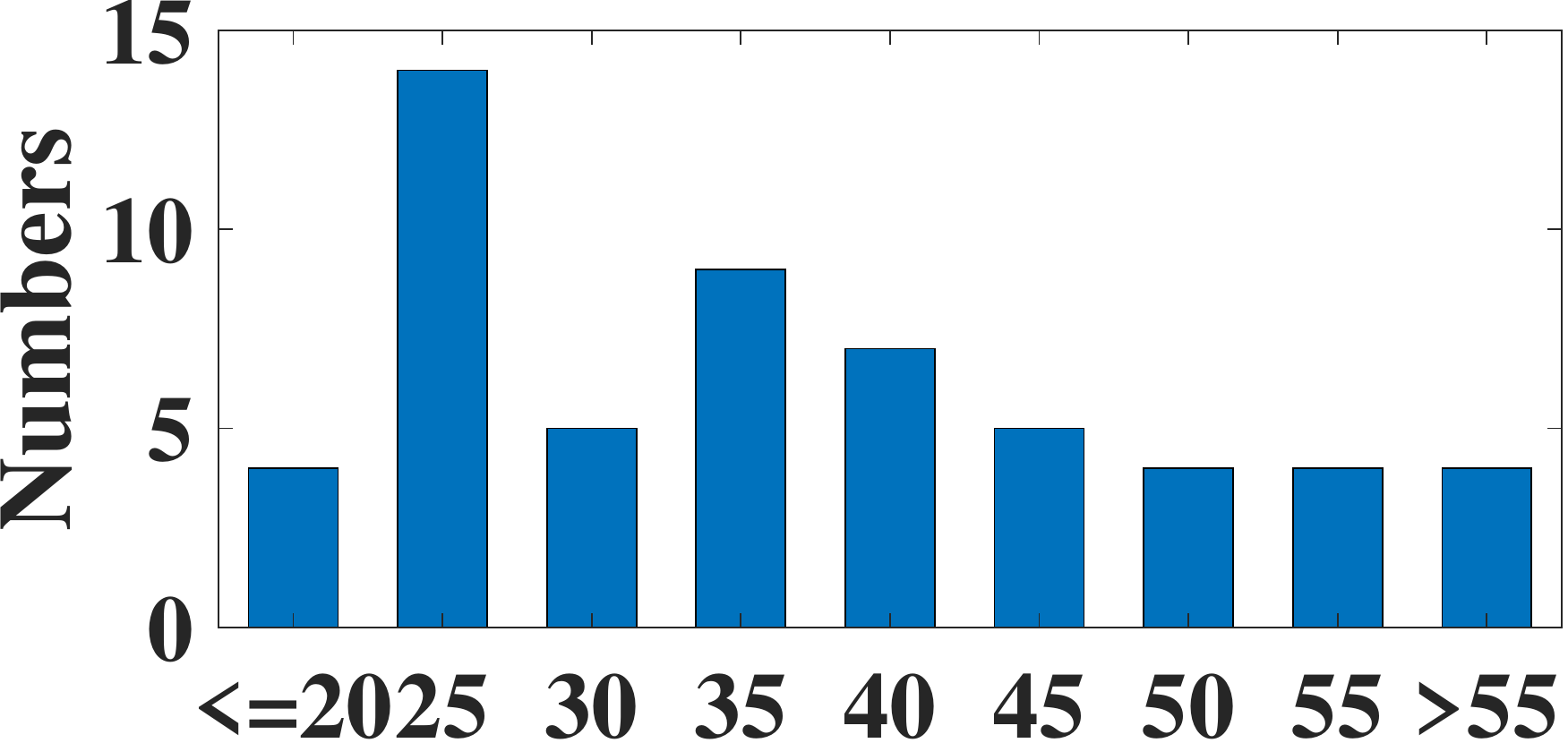}  
            \vspace{-7mm}
            \label{fig:age}
        \end{minipage}
    }
    \subfigure[Height groups.]
    {
        \begin{minipage}[t]{0.48\linewidth}
            \centering
            \includegraphics[width=0.8\linewidth]{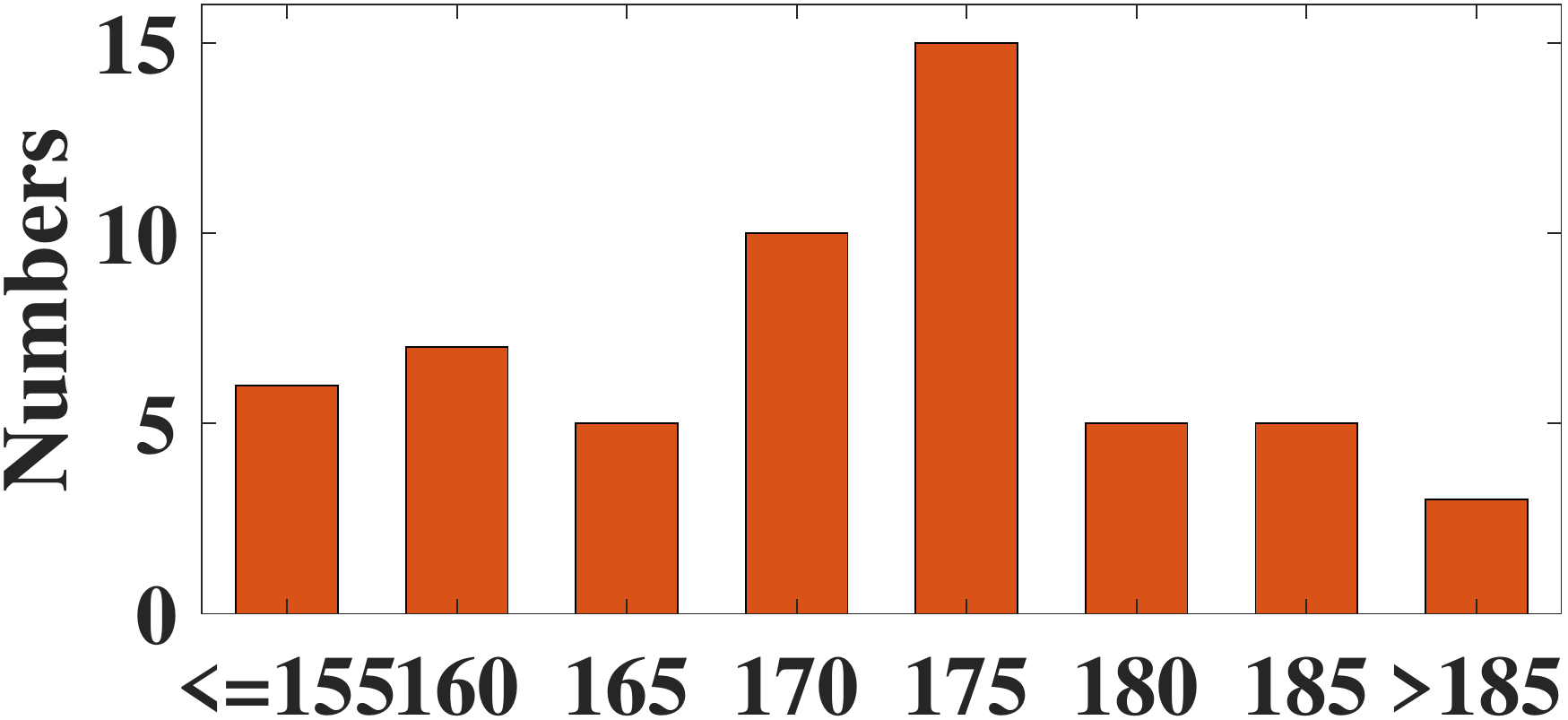}
            \label{fig:height}
            \vspace{-7mm}
        \end{minipage}
    }\vspace{-4mm}
    \caption{The cohort stats regarding the age and height. (a) Age groups. (b) Height groups.}
    \vspace{-3mm}
    \label{fig:stats}
\end{figure}

\textcolor{black}{
We recruited a total of 56 participants to collect mmWave and vision data. The participants consist of 26 males and 30 females, aging from 17 to 60 with various heights from 152cm to 188cm and weights from 45kg to 96kg. Fig.\ref{fig:stats} shows the cohort stats regarding the age and height, which are generally believed to have significant influence on gait patterns. As for the details on the age of subjects, we divide all subjects into small groups every five years and the boundaries are open at the front and closed at the back (i.g., ( , ]). For example, the 30 of abscissa axis in Fig.\ref{fig:age} represents (25,30]. 
We also divide all subjects into small groups every 0.05m on the height.
The cohort stats of the participates depicted in Fig.\ref{fig:stats} demonstrates that the participates in our evaluation have diversified ages and body characteristic (e.g., heights).
}

\subsection{Data Collection}

{\bfseries mmWave Radar Platform.} For the radar data collection, we utilize a commercial and off-the-shelf millimeter-wave radar IWR6843-BOOST. The radar operates in a frequency band from 60GHz to 64GHz whose wavelength is $\sim$ 4mm. It has three transmitting antennas and four receiving antennas that form a 60 degree azimuth FoV and 60 degree elevation FoV whose angle resolution is $\sim15^{\circ}$. We utilize the FMCW processing chain provided by TI and the radar outputs 3D point cloud. \textcolor{black}{ For reproduction, the detailed configuration parameters of the device are provided as follows: 
the device is set to transmit 32 chirps per frame. The start frequency of the chirp is set to 60.065GHz. The frequency bandwidth is set to 3194.88MHz. The Frequency slope is set to be 12.5MHz/us. }



\begin{figure*}[h]\vspace{-0.5mm}
  \centering
  \includegraphics[width=0.9\textwidth]{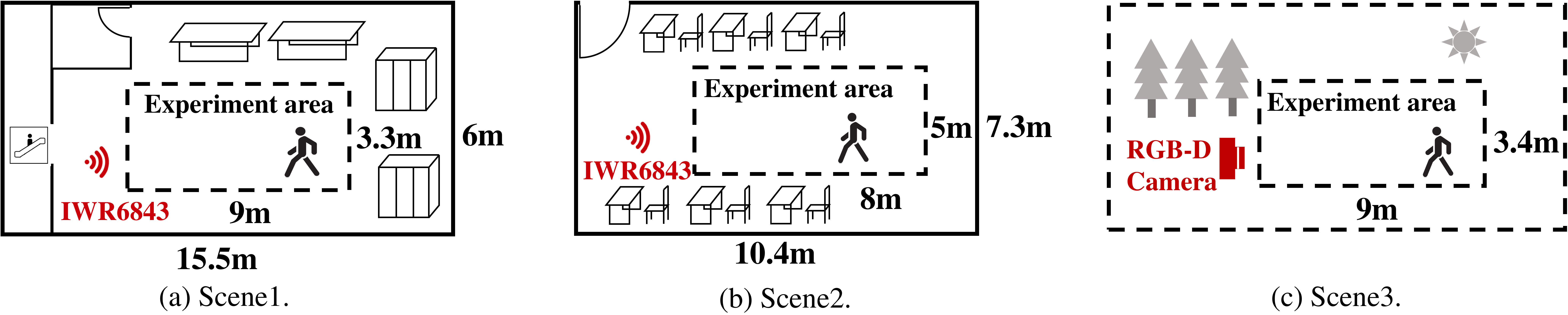}
  \vspace{-2mm}
  \caption{Experimental scenarios. (a), (b)MmWave data collection areas indoors, simulating the \textcolor{black}{camera-restricted}  scenario. (c) Vision data collection area outdoors, simulating the \textcolor{black}{camera-allowed} scenario.}
  \label{fig:experiments}
  \vspace{-3.5mm}
\end{figure*}

\noindent \textbf{Experiment Site and Data Collection.} For experiments, we collect radar data in two scenes: a corridor and a room. As shown in Fig.\ref{fig:experiments} (a) and (b), the millimeter-wave device is placed at one end of the corridor or room with a height of 0.9m, and participants on the other end walk towards the radar device. \textcolor{black}{ For multi-person scenario, the closest distance between two concurrent people is about 0.3m which is a reasonable social distance as adopted similarily in the relevant studies \cite{yang2020mu, meng2020gait}}. In each experiment, a participant walks for at least $7$ meters. Note that because of different people's walk speeds, the recording duration of a radar capture is different and fluctuates ranging from 4.5s to 6.5s. In order to create a valuable data set and avoid data corruption, participants are required to walk ten times or more. In summary, the dataset contains 45 different identities with each identity having 17 radar records on average. It took about 30 days to collect the data and people were free to change clothes across during this period.

\noindent \textbf{RGB-D Camera and Experiment.}
To collect the dense mesh of people, we utilize Azure Kinect which is a representative type of RGB-D camera. As shown in Fig.\ref{fig:experiments} (c), the RGB-D camera is placed at one end of the experiment area with a height of 1m. 
The sites for camera data collection are completely disjoint from the locations used for radar data collection, mimicking the real-world public and privacy-private scenarios respectively. We collected the camera data of 45 participants walking in the entrance area and outdoor environments. Each participant was required to walk a total of 20 times. Finally, we convert the camera data into the same Cartesian coordinate system as the radar data.

\subsection{Preprocessing for Radar Data}

While the human subjects are easy to segment from RGB-D camera data, the segmentation with radar points is much more tricky. The radar data was collected in confined indoor environments that include static obstructions such as walls, floors, and ceilings, together with the walking people. In such an environment, however, multi-path noise is non-negligible, which is a common issue for almost all RF technologies. Due to the reflection of ambient objects and beam spreading \cite{lu2020milliego}, the propagation of mmWave signals between objects and transceivers tends to travel through multiple paths. Consequently, unwanted points often appear in the radar point cloud which are widely known as the 'ghost points' \cite{lu2020see}. In order to mitigate the impact of these noisy points and segment human subjects out, we implement the following two steps (1) Height Heuristics based Denoising (2) Clustering-based Point Segmentation.
 
\subsubsection{Denoise with Height Heuristics.}  we first remove some unlikely points based on their 3D locations as we know the normal areas where people walk. For instance, when the device is placed 1m high above the ground and the coordinate system is based on the device as the origin, points with their heights greater than 2.5m or below than -1m are unlikely to be human bodies that can be safely removed.
 
\subsubsection{Segmentation via Clustering.}
Next we apply the DBScan algorithm \cite{birant2007st} to acquire the cluster of points of a person such that the near-person noise can be suppressed. DBScan is a density-aware clustering algorithm that can divide a point cloud based on the distance and the density described based on a set of neighborhoods in the 3D space.  As it does not require the number of clusters to be specified a priori and can automatically mark outliers that are noise, DBScan has been utilized by \cite{zhao2019mid} to separate individual human objects from mmWave radar point clouds. Our implementation carefully follows \cite{zhao2019mid} to separate the radar points belonging to different individuals. Regarding the hyperparameters settings of DBScan, we empirically set the maximum distance (radius) between two points falling into the same cluster to $0.35$ and set the minimum point number in a cluster to $3$. 

\begin{figure}[h]\vspace{-2.5mm}
    \centering
    \subfigure[Single subject.]
    {
        \begin{minipage}[t]{0.40\linewidth}
            \centering
            \includegraphics[width=0.8\linewidth]{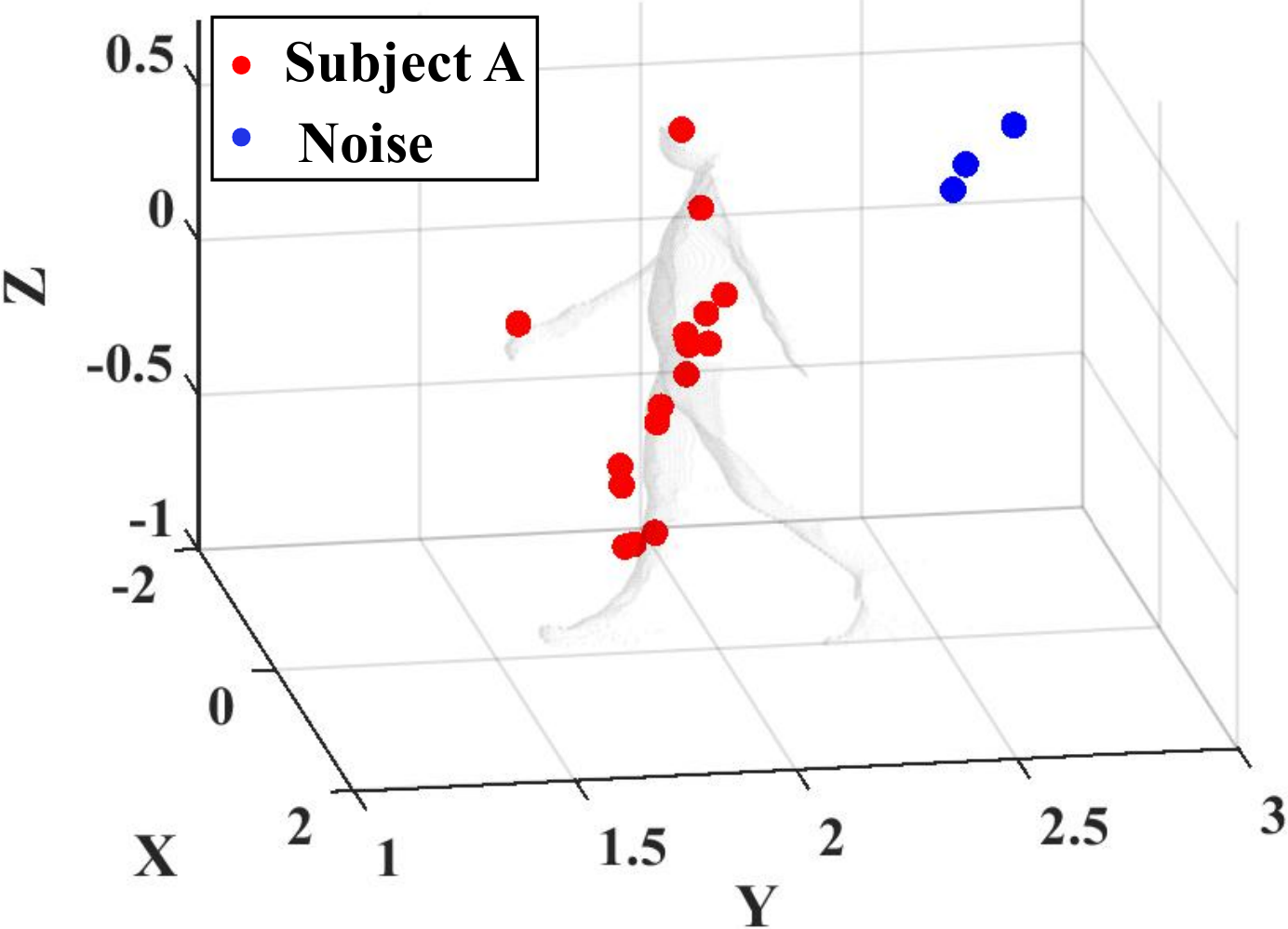}  
            \label{}
            \vspace{-9mm}
        \end{minipage}
        \vspace{-9mm}
    }
    \subfigure[Two subjects.]
    {
        \begin{minipage}[t]{0.40\linewidth}
            \centering
            \includegraphics[width=0.8\linewidth]{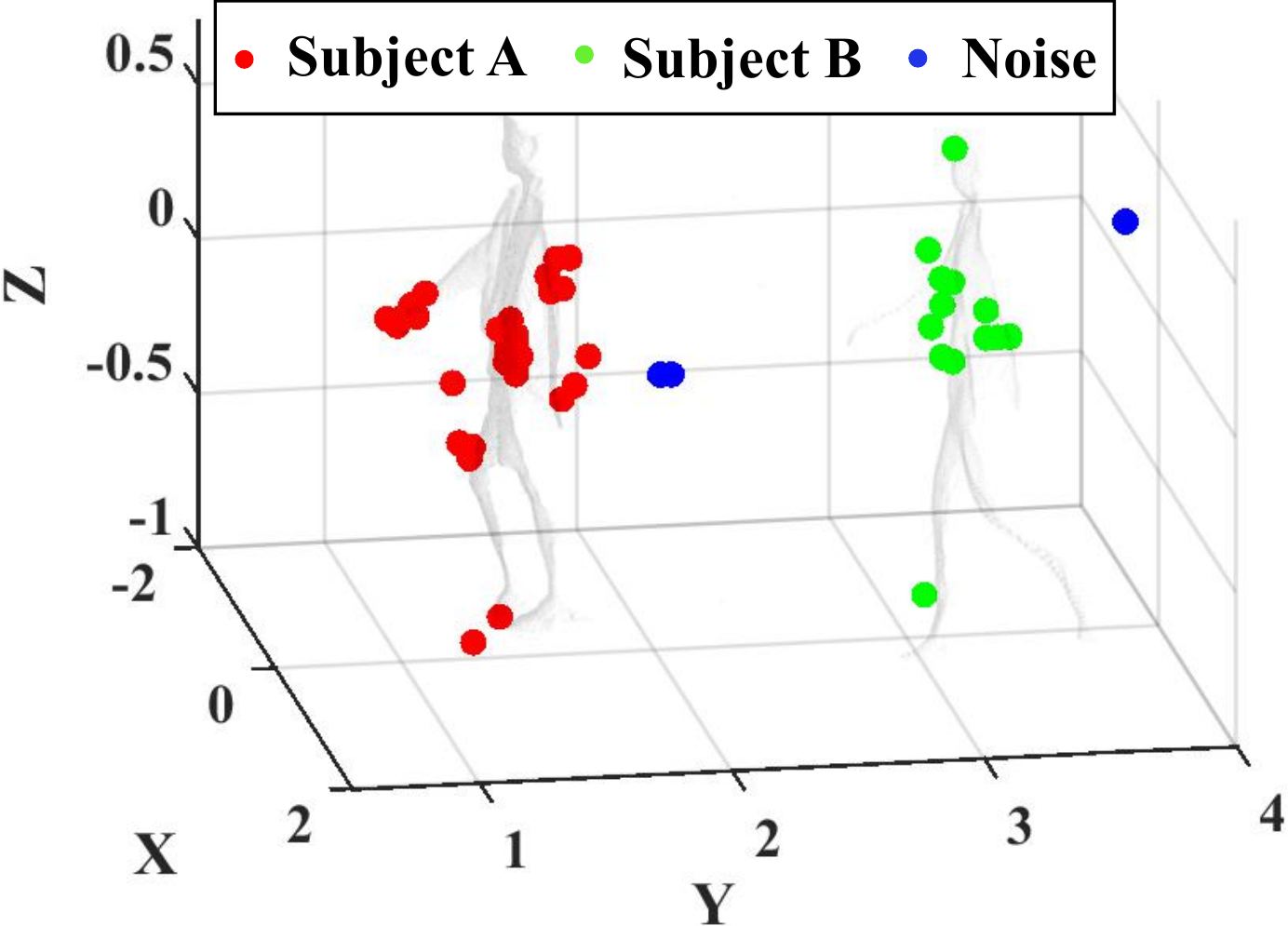}
            \label{}
            \vspace{-9mm}
        \end{minipage}
        \vspace{-9mm}
    }
    \vspace{-3.5mm}
    \caption{Segmented point clouds belonging to individual people using height heuristics and clustering.}
    \label{fig:human_cluster}
    \vspace{-3mm}
\end{figure}

As shown in Fig.\ref{fig:human_cluster}, after applying the two denoising steps, radar points belonging to two different individuals can be clearly segmented.




%% file: 9Performance_Evaluation.tex
\UseRawInputEncoding
\vspace{0.5mm}
\section{System Evaluation}\label{sec:evaluation}
 
This section presents the performance evaluation of our system. We start with the evaluation methodology including the training and test procedure, evaluation metrics and competing approaches (Section \ref{sec:methodology}). The overall performance is reported in Section \ref{sec:overall}, followed by separate evaluations of each critical design component in Section \ref{sec:components}. We then extensively study the impact of various factors on the performance in Section \ref{sec:sensitivity}.
\vspace{-1.5mm}
\subsection{Evaluation Methodology}\label{sec:methodology}
As introduced in Section \ref{sec:implementation}, we establish the databases for radar and RGB-D data of 56 subjects. To evaluate the identification accuracy for both single-person presence in the scene and the co-presence of multiple people at the same time, a single-person database and a multi-person database are respectively built. The former contains 655 radar records with one person in the FoV, while the latter consists of 248 records with 2 people in the FoV. In addition, to evaluate the impact of the angle of view, we collect a total of 1600 radar records with varying angles. On the other hand, the RGB-D gallery contains 1072 records in total. We have up to 17 RGB-D records for each subject, representing the real-world situation where people are typically captured by camera several times. Also, as each gait cycle typically spans about 1s, all records of both modalities contain the sensing data of several gait cycles. \textit{For reproducibility, this multi-modal dataset used in our evaluation will be released to the community.}

{\color{black}
\subsubsection{Model Setting and Model Training/Testing}

In this section, we describe the details of the deep metric learning model, training, and testing procedure. In pointNet, we implement a multi-layer perception (MLP), the layer sizes of which are $(x,12,24,48,64)$ with $x$ being 5 and 4 for mmWave points and RGB-D signature respectively. We use batch normalization followed by ReLU activation functions after all layers. The attention operation is implemented by fully connected layer (64,1). The LSTM has 3 layers and the size of each layer is 64.

Following the conventional strategy adopted by many cross-modal ReID studies in computer vision \cite{haque2016recurrent, karianakis2018reinforced},  we use $75\%$ of the data records collected from each subject for model training, and the remaining $25\%$ serve as the testing set. The learning rate is set to 0.0002 and the batch size is 32. The number of training epochs is 50000. The hyper-parameter assigned to loss function in Equation \ref{equ:loss} (i.e., $margin$) is set to 0.3. We implement our deep learning model in PyTorch and  train the model with NVIDIA RTX 3090.

In each ReID test, we randomly select one radar record from test set as the query and our system retrieves the RGB-D record of the same identity from the RGB-D gallery. Note that these query and gallery records are not used in the training phase. Single-person ReID and multi-person ReID experiments are conducted using radar records with one and two subjects respectively.  To make the result more accurate, we perform 4-fold cross-validation, where each time we randomly assign records to the training set and test set.
}

 
\subsubsection{Evaluation Metrics}  Our evaluation  follows \cite{li2018harmonious} and adopts the cumulative matching curve (CMC), which is a widely adopted metric in ReID studies. Specifically, for each test sample,  we calculate the similarity between the radar query and every candidate's RGB-D record. The results are ranked and the top-N RGB-D records (the N most similar records) are retrieved. We report the top-N accuracy, which is defined as the percentage of test cases where the RGB-D record of the target person is ranked among the top N positions among all the RGB-D records in the test. N varies from 1 to 9 in our evaluation given the number of our volunteers.

{\color{black}
\subsubsection{Competing Approaches}
We compare our approach with the following four baselines. Since this is the first work for cross-modal ReID among vision and radar, we port several single-modal mmWave or RGB-D (Re)ID approaches in the recent literature to our problem. 

\vspace{1mm}
\noindent {\bfseries PointNet + LSTM (PL)} \cite{qi2017pointnet,cheng2021person}. We adapt the single-modal mmWave points re-identification method \cite{cheng2021person} to our problem. Specifically, the method first uses PointNet \cite{qi2017pointnet} to extract features from dense points from raw RGB-D output as well as  raw mmWave points. Secondly, the feature vector is fed to LSTM to extract temporal and spatial features. Lastly, a mean pooling operation aggregates the features to get the final gait embedding. The training method is also based on triplet loss.

\vspace{1mm}
\noindent {\bfseries Voxelization + 3DCNN + LSTM (VCL)} \cite{zhao2019mid}. We also implement mmWave person identification method \cite{zhao2019mid} based on voxelization and 3DCNN. In this baseline method, a point cloud is first mapped to a 3D voxel grid, and then the 3D voxel grid is converted into a feature vector using 3DCNN. Finally, LSTM and mean pooling are utilized to obtain the gait embedding.
\vspace{1mm}
\noindent {\bfseries DGCNN + LSTM (DGL)} \cite{zheng2020person,wang2019dynamic}. We port the single-modal RGB-D person ReID method \cite{zheng2020person} to our problem. Specifically, the method utilizes DGCNN \cite{wang2019dynamic}, a graph CNN to extract features from raw RGB-D outputs and mmWave points. DGCNN consumes the point cloud directly and applies the proposed EdgeConv which takes $k$ adjacent points as graph structure to extract local features. Finally, LSTM and mean pooling are utilized in the same way as VCL to extract gait embedding.
\vspace{1mm}
\noindent \textbf{Earth Mover’s Distance (EMD)} \cite{rubner1998metric}. This baseline utilizes the traditional similarity estimation method (i.e., EMD) to measure the similarity between a mmWave capture and a synthesized RGB-D signature.  Earth Mover’s Distance (EMD) \cite{rubner1998metric} is a popular method to qualify the distance between two distributions. In short, it models two distributions as a mass of earth and a collection of holes and it measures the least amount of work needed to fill the holes with earth. In our problem, both the synthesized signature points and radar points essentially capture the distribution of specular reflection energy in the space over time. Therefore, their similarity can be estimated by calculating the distance between two distributions.

}





\subsection{Overall ReID Performance} \label{sec:overall}

We now report the overall ReID accuracy of single-person (i.e., only one subject appears in the radar FoV) and multi-person (i.e., more than one subjects appears in the radar FoV)  experiments.
 
\color{black}
\vspace{-1.5mm}
\subsubsection{Single-person ReID accuracy} 
\begin{figure}[t]\vspace{-7mm}
    \centering
    \subfigure[Single subject.]
    {
        \begin{minipage}[t]{0.45\linewidth}
            \centering
            \includegraphics[width=0.96\linewidth]{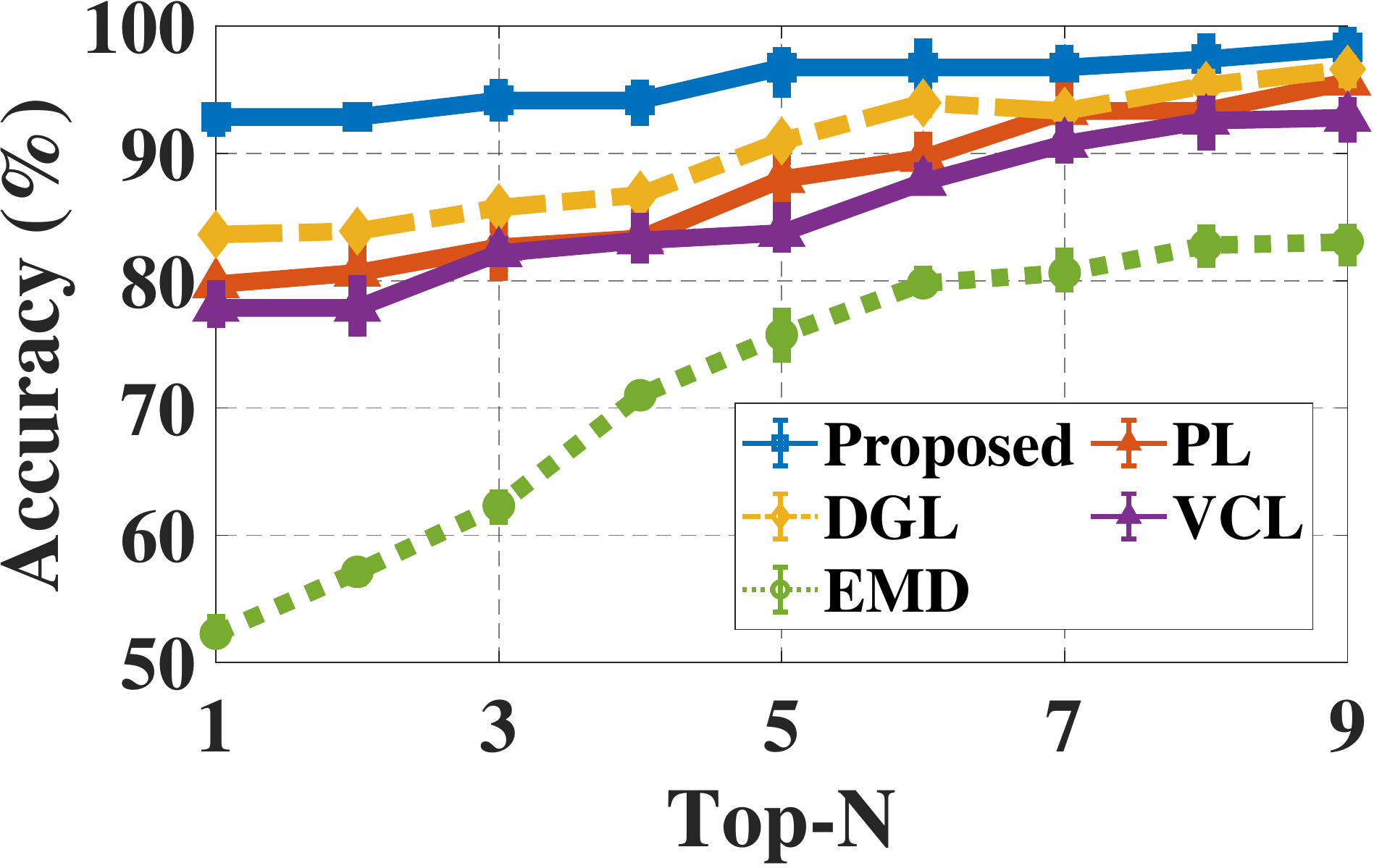}  
            \vspace{-8mm}
            \label{}
        \end{minipage}
    }
    \subfigure[Multiple subjects.]
    {
        \begin{minipage}[t]{0.45\linewidth}
            \centering
            \includegraphics[width=0.96\linewidth]{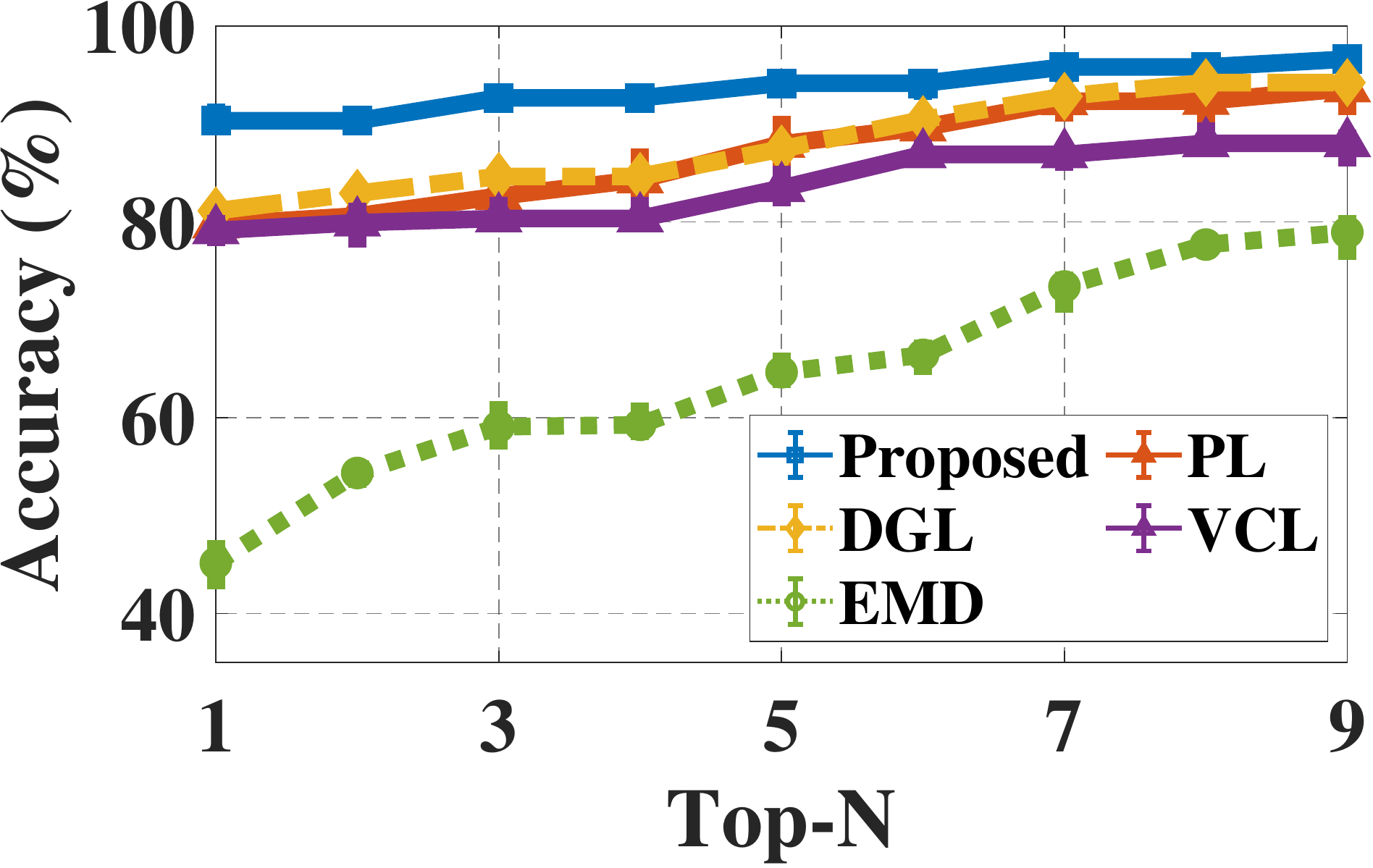}
            \label{}
            \vspace{-8mm}
        \end{minipage}
    }\vspace{-5mm}
    \caption{Overall performance of ReID (Cumulative Matching Curve). (a) Single-person ReID. (b) Multi-person ReID.}
    
    \label{sec:overallperformance}
    \vspace{-5mm}
\end{figure}

{\color{black}
As Fig.\ref{sec:overallperformance}(a) depicts, our design achieves $92.86\%$ top-1 accuracy, $94.16\%$ top-3, and $96.75\%$ top-5 accuracy out of 56 subjects in the single-person ReID. In other words, given the radar capture of a person of interest, we rank 56 candidates' RGB-D meshes based on cross-modal similarity. We have a $92.86\%$ chance to identify the target subject as the most similar candidate. The chances to have the target among the most similar 3 and 5 candidates are $94.16\%$ and $96.75\%$ respectively.

The performance of our method significantly outperforms the baseline approaches. First, our approach achieves $13\%$ higher top-1 and $7\%$ higher top-5 accuracy than the \textbf{PL}  using raw RGB-D data (i.e., without being preprocessed by specular reflection model), showing that the importance of the signature synthesis (Section \ref{sec:signature_synthesis}) in addressing modality discrepancy.
Our method also outperforms other two methods ported from single-modal re-identification (i.e., \textbf{VCL} and \textbf{DGL}) especially for top-1 accuracy, improving by at least $9\%$. 
Second, compared with the traditional similarity estimator (i.e., \textbf{EMD}), our method demonstrates dramatic improvement (e.g., improving for top-1 accuracy by 39\%) indicating that our similarity estimation algorithm using deep metric learning more robust against interference factors (e.g., temporal misalignment and minor variation in gaits). 
Finally, we also compare our method with an  unsupervised learning method for cross-modal ReID. Specifically, we utilize representative methods (i.e., PointNet \cite{qi2017pointnet} and RNN) to extract features of internal and inter frames for radar and RGB-D point cloud, and multi-label classification loss \cite{wang2020unsupervised} as the loss function. However, because of the manifestation gaps among heterogeneous sensor modalities, we found that the pseudo labels predicted by the unsupervised method are very error-prone on our cross-modal Re-ID problem. The top-5 accuracy of unsupervised learning evaluation on our cross-modal task is only $40\%$ on average. This indicates that supervised learning is still a better learning fashion over unsupervised learning for cross-modal Re-ID. }


\color{black}
\vspace{-1.5mm}
\subsubsection{Multi-person ReID accuracy}
{\color{black}
Fig.\ref{sec:overallperformance}(b) shows that our method achieves a comparable accuracy in the multi-person ReID with $90.33\%$ top-1 accuracy, $92.67\%$  top-3, and $94.16\%$ top-5 accuracy. This result validates that our method is very robust even when there is more than one subject co-present in the radar FoV, demonstrating the unique advantage of our design against the WiFi-based solution \cite{korany2019xmodal} in the capability of simultaneously handling multiple people of interest in the real-world settings.}

\color{black}
\vspace{-1mm}
\subsection{Effectiveness of Key Design Components}\label{sec:components}

\subsubsection{Ablation Study}
In Section \ref{sec:design}, several critical designs were introduced. To demonstrate their effectiveness, we measure the performance of the system when specific components are disabled. We conduct both single-person and multi-person ReID experiments with the following 4 different settings and top-5 accuracy is provided.

\color{black}
\noindent \textbf{Full version (Full):} All the components in Section \ref{sec:design} are activated.

\noindent \textbf{w/o spatio-temporal signature (noST):} The similarity is computed on raw input data. The RGB-D mesh is not processed by specular reflection model (Section \ref{sec:signature_synthesis}) to synthesize signature. This setting aims to highlight the importance of using signature points rather than raw inputs.

\noindent \textbf{w/o attention (noAtt):} The attention operation in similarity estimation module (Section \ref{sec:similarity}) is replaced with max-pooling operation utilized in the original PointNet paper. This setting aims at examining effectiveness using the attention operation to our problem (e.g., dynamically adjust the weights of various body parts).

\noindent \textbf{w/o triplet loss (noTL):} The training strategy is replaced by two-tuples inputs with contrastive loss. In specific, the network is trained with two-tuples consisting of one sample from each modality (denoted by $<A,B>$) and the loss function is the contrastive loss \cite{ge2018deep} defined as:
\begin{equation}
 L_c = yd^2+(1-y)max(margin_c-d,0)^2
 \label{equ:contrastive}
\end{equation}
where $d$ represents the distance of $A$ and $B$.   $margin_c$ is a pre-defined threshold. $y$ is the label of of the two samples, indicating whether A and B from two modalities belong to the same identity ($y = 1$) or not ($y = 0$). This setting aims to evaluate the importance of the triplet loss during training stage to the performance (Section \ref{sec:triplet}).

\begin{figure}[t]\vspace{-9mm}
    \makeatletter\def\@captype{figure}\makeatother
    \begin{minipage}[h]{0.45\linewidth}
            \flushleft
            \includegraphics[width=1\columnwidth]{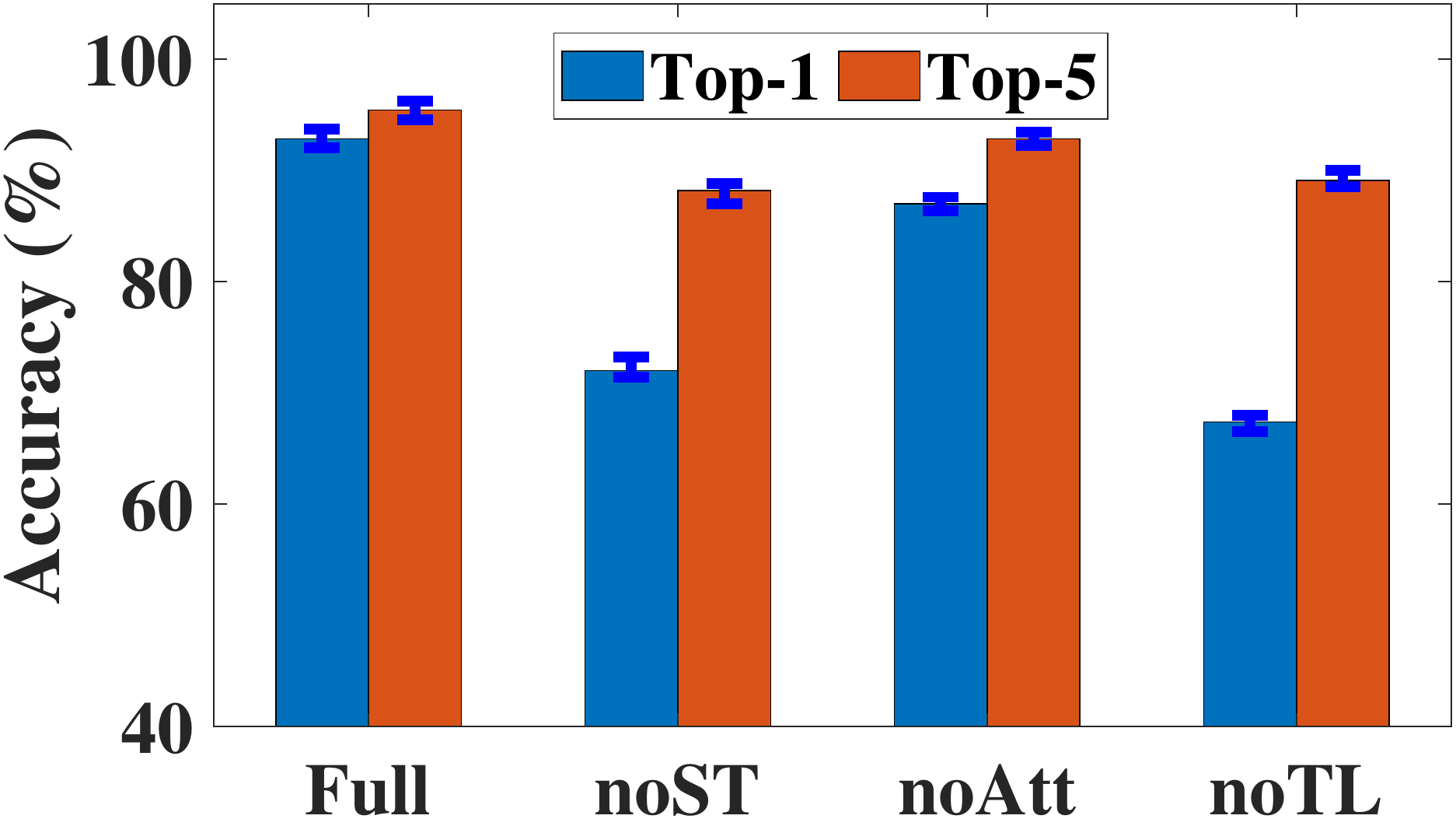}
            \vspace{-6.5mm}
            \caption{Effectiveness of diff design components. \quad }
            \Description{}
            \label{fig:effectiveness}
    \end{minipage}
    \makeatletter\def\@captype{figure}\makeatother
    \begin{minipage}[h]{0.48\linewidth}
        \centering
        \vspace{0.5mm}
        \includegraphics[width=0.8\linewidth]{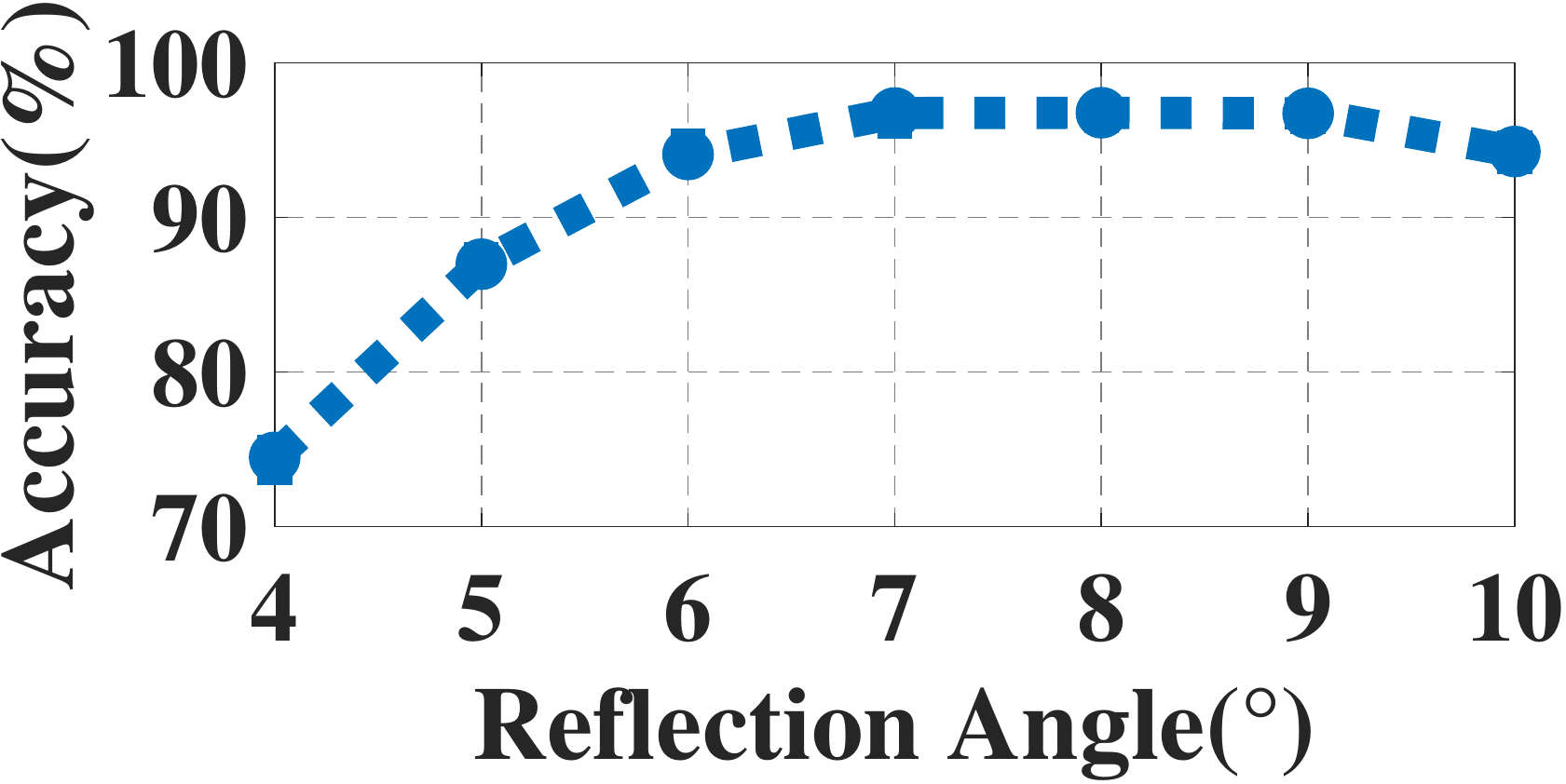}\vspace{-1.5mm}
        \caption{Impact of reflection angle threshold.}
        \label{fig:angle}
        \vspace{-2mm}
    \end{minipage}
     \vspace{-4.5mm}
\end{figure}





The results are shown in Fig.\ref{fig:effectiveness}. The full version setting achieves the best results, confirming that every design component is effective to cross-modal ReID. Furthermore, compared with the results of \textbf{noST}, we observe that the signature points plays an very important role in our design. We therefore hypothesize that spatio-temporal signature points preserve the most salient characteristic of an identity while unifying the modality representation. Moreover, it eliminates the the redundancy in raw RGB-D data, and thus is easier for the network to extract human gait features. The result of \textbf{noAtt} demonstrates that attention operation also effectively improves the accuracy while the improvement on top-1 is most significant. This is because by adopting the attention mechanism, we can exploit our observation in Fig.\ref{fig:signatureDiff} - the network paies more attention to features on the limb parts which are more identity-specific  than the torso that has more points due to larger RCS but contains limited identify information.
Finally, the performance drop shown in \textbf{noTL} proves that the triplet loss is more effective than contrastive loss.

\color{black}
\vspace{-2mm}
\subsubsection{Reflection angle threshold}
In signature synthesis (Section \ref{sec:signature_synthesis}), we utilize a parameter  $\epsilon$ as the threshold to determine signature points. An improper setting of $\epsilon$ might lead to incorrect signature points, whereas there will be fewer points if $\epsilon$ is too small. To find the optimal threshold, we examine the performance over various threshold values. {\color{black} As Fig.\ref{fig:angle} depicts, the optimal value of $\epsilon$ is empirically found at $7^{\circ}$.} 

           

\color{black}
\vspace{-2mm}
\subsection{Sensitivity Analysis}\label{sec:sensitivity}

\vspace{-1mm}
\subsubsection{Impact of view angles.} Our approach can work in real-world scenarios where  RGB-D and radar sensors are installed in disjointed locations and thus have different view angles of a walking subject. This mainly benefits from our body mesh reconstruction (discussed in \ref{sec:signature_synthesis}) on RGB-D data and rich spatial information available from radar. We therefore evaluate the performance against various view angles of both sensors.  

\begin{figure}[b]\vspace{-5.8mm}
    \centering
    \subfigure[Various view angles of RGB-D camera.]
    {
        \begin{minipage}[t]{0.45\linewidth}
           \centering
           \hspace{-3mm}
           \includegraphics[width=1\linewidth]{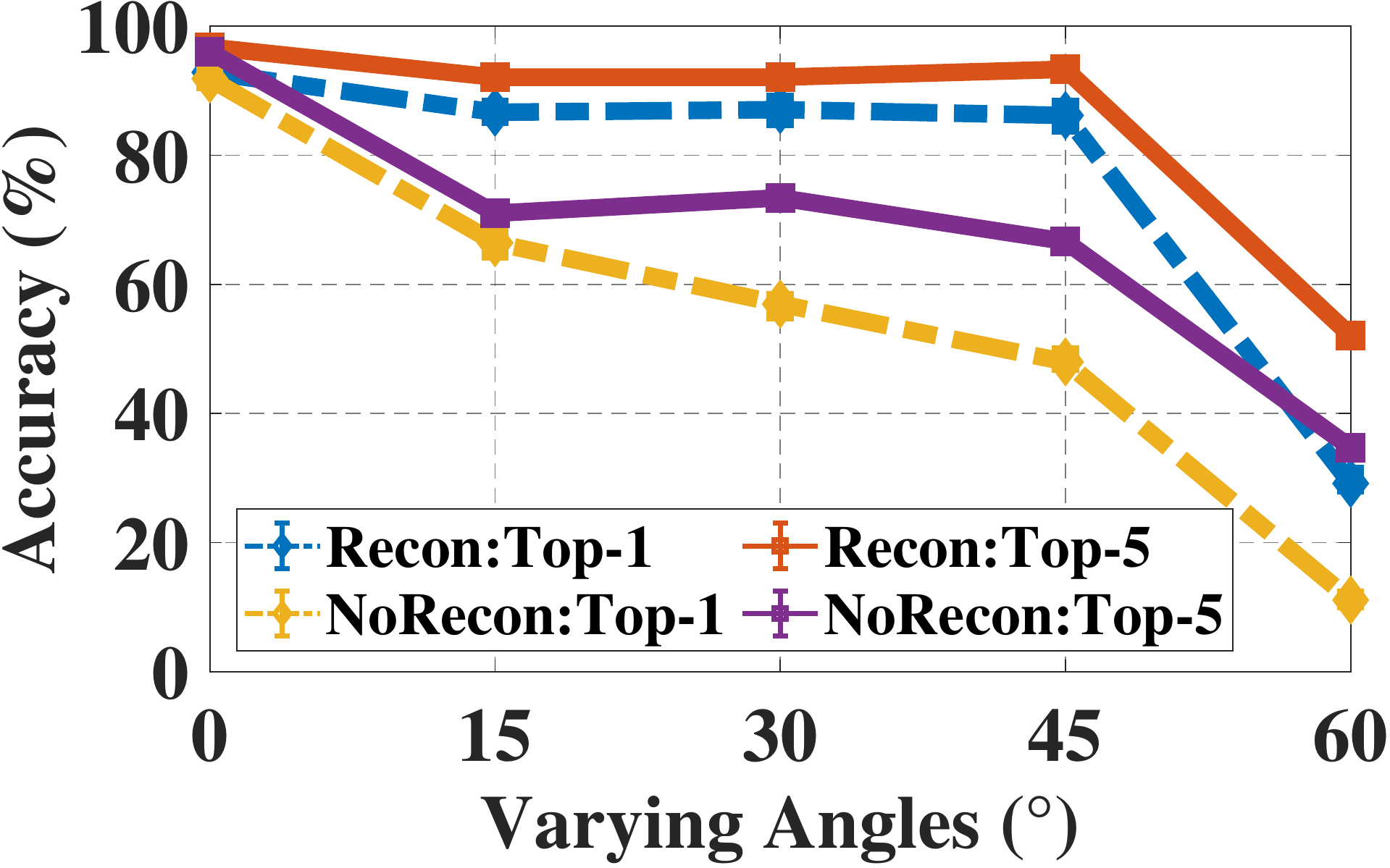}
           \label{fig:crossangle}
          \vspace{-4mm}
        \end{minipage}
    }
    \subfigure[Various view angles of radar.]
    {
        \begin{minipage}[t]{.45\linewidth}
           \flushleft
           
           \includegraphics[width=1\linewidth]{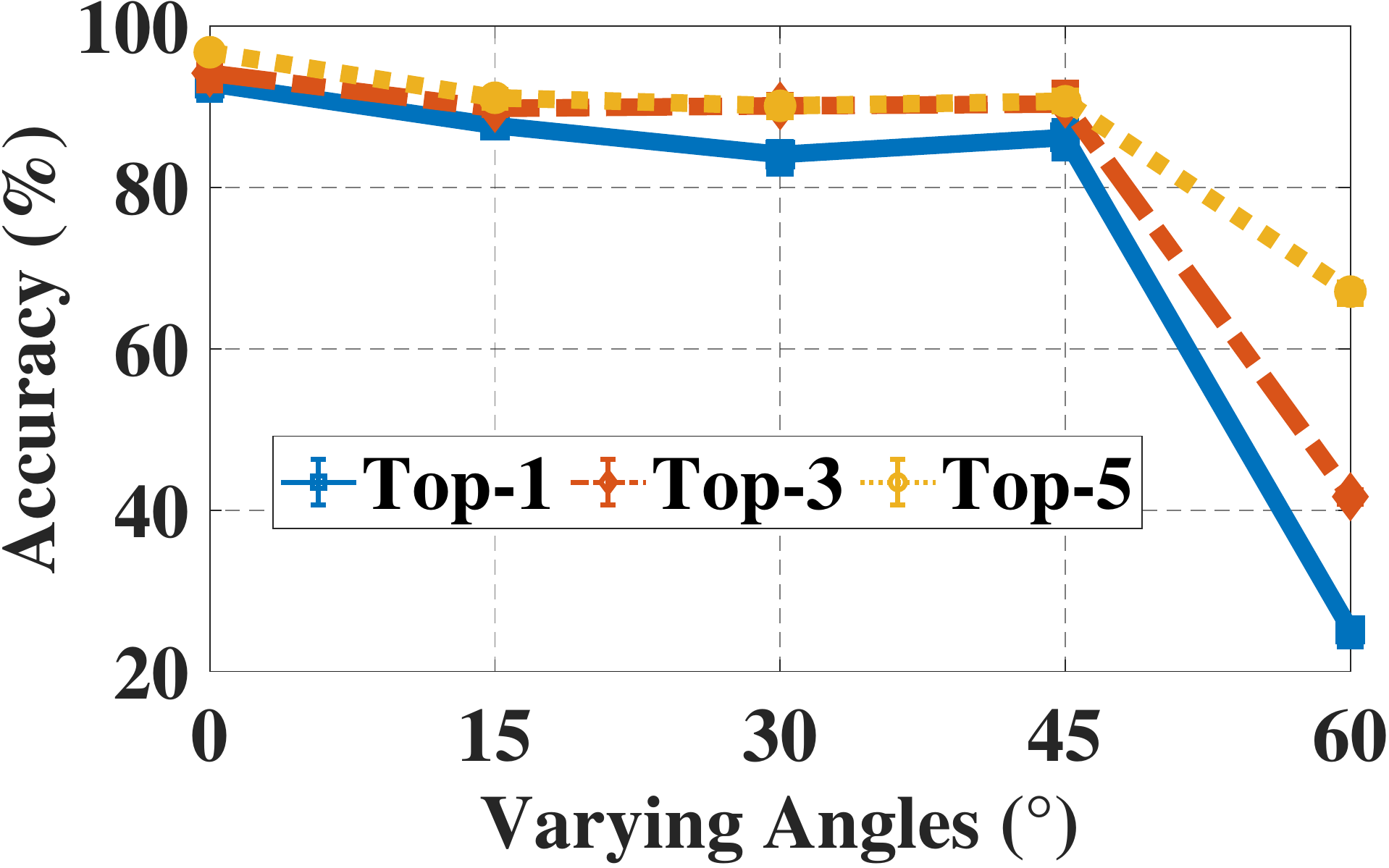}
           \label{fig:crossanglemm}
          \vspace{-4mm}
        \end{minipage}
    }
    \vspace{-4.5mm}
    \caption{Evaluation of various view angles.}
    \label{fig:variousAngles}
    \vspace{-5mm}
\end{figure}

\noindent \textbf{RGB-D camera view angles.} To simulate different view angles of the camera, candidates are asked to walk following a series of trajectories with varying offsets (from 0 to 1.2 meters) from the camera's mid-line (i.e., 0 meter). {\color{black}
By doing this, the angle between the target and the device varies from $0^{\circ}$ to $60^{\circ}$ with the interval of $15^{\circ}$. On the other hand, the target subject walks towards the mmWave radar along the mid-line that is perpendicular to the radar. This simulates common RF sensing scenarios (e.g., walking in camera-restricted areas such as corridor) and it also allows the radar to perceive as much gait information as possible. The results in Fig.\ref{fig:crossangle} demonstrate that our system shows stable top-1 accuracy ($86.18\%$) and top-5 accuracy ($93.3\%$) at a large angle of $45^{\circ}$ (0.9 meters offset), while accuracy drops dramatically when the angle is greater than $60^{\circ}$ because subjects are out of the FoV of the camera and the data captured is incomplete. To further demonstrate the benefits of mesh reconstruction in mitigating view angle issues (e.g., self-occlusion), Fig.\ref{fig:crossangle}  compares accuracy with and without performing reconstructions. The reconstruction improves both top-1 and top-5 accuracy by up to $26\%$ at large view angles (i.e., $45^{\circ}$), showing its effectiveness in mitigating self occlusions.
}

\noindent\textbf{mmWave radar view angles.} We also measure the impact of radar view angles. In the experiments, the target subject walks toward the radar following the trajectories with varying offsets from the mid-line (from 0 to 1.2 meters).
{\color{black} The angle between the subject and the radar varies from $0^{\circ}$ to $60^{\circ}$ with the interval of $15^{\circ}$ to simulate different view angles. The results depicted in Fig.\ref{fig:crossanglemm} demonstrates that robust top-1 ($> 86\%$) and top-3 ($> 90\%$) accuracy across various view angles from $0^{\circ}$ to $45^{\circ}$ (from 0 to 0.9 meters offset). The problem becomes more difficult when the angle between subject and radar is greater than at $60^{\circ}$ due to two challenges. First, the angular resolution and signal strength decrease dramatically at large angle. Second,  identify-specific signature (e.g., arm swings) may be lost due to self-occlusion, making the signature less distinguishable. The result suggests that we can install multiple radar sensors to provide various view angles, which can be combined to improve the accuracy.}

\vspace{-2mm}
\subsubsection{Performance of subject walking away from the device.}
\begin{figure}[b]\vspace{-7mm}
    \makeatletter\def\@captype{figure}\makeatother
    \begin{minipage}[h]{0.48\linewidth}
            \flushleft
            \includegraphics[width=0.9\columnwidth]{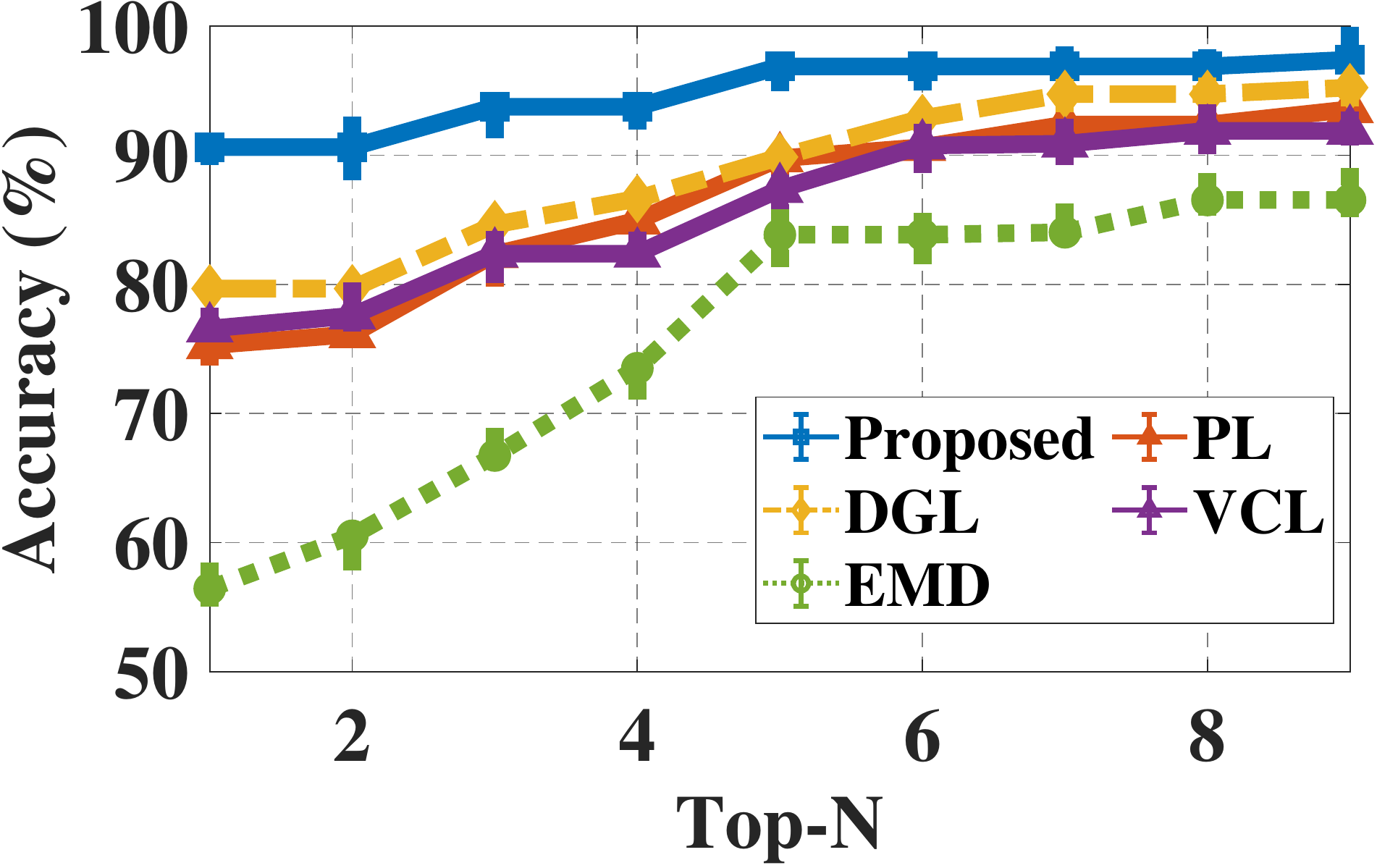}
            \vspace{-3mm}
            \caption{Performance of subject walking away from device. \quad }
            \label{fig:walkingaway}
            \vspace{-2mm}
    \end{minipage}
    \makeatletter\def\@captype{figure}\makeatother
    \begin{minipage}[h]{0.48\linewidth}
        \centering
        \includegraphics[width=.9\linewidth]{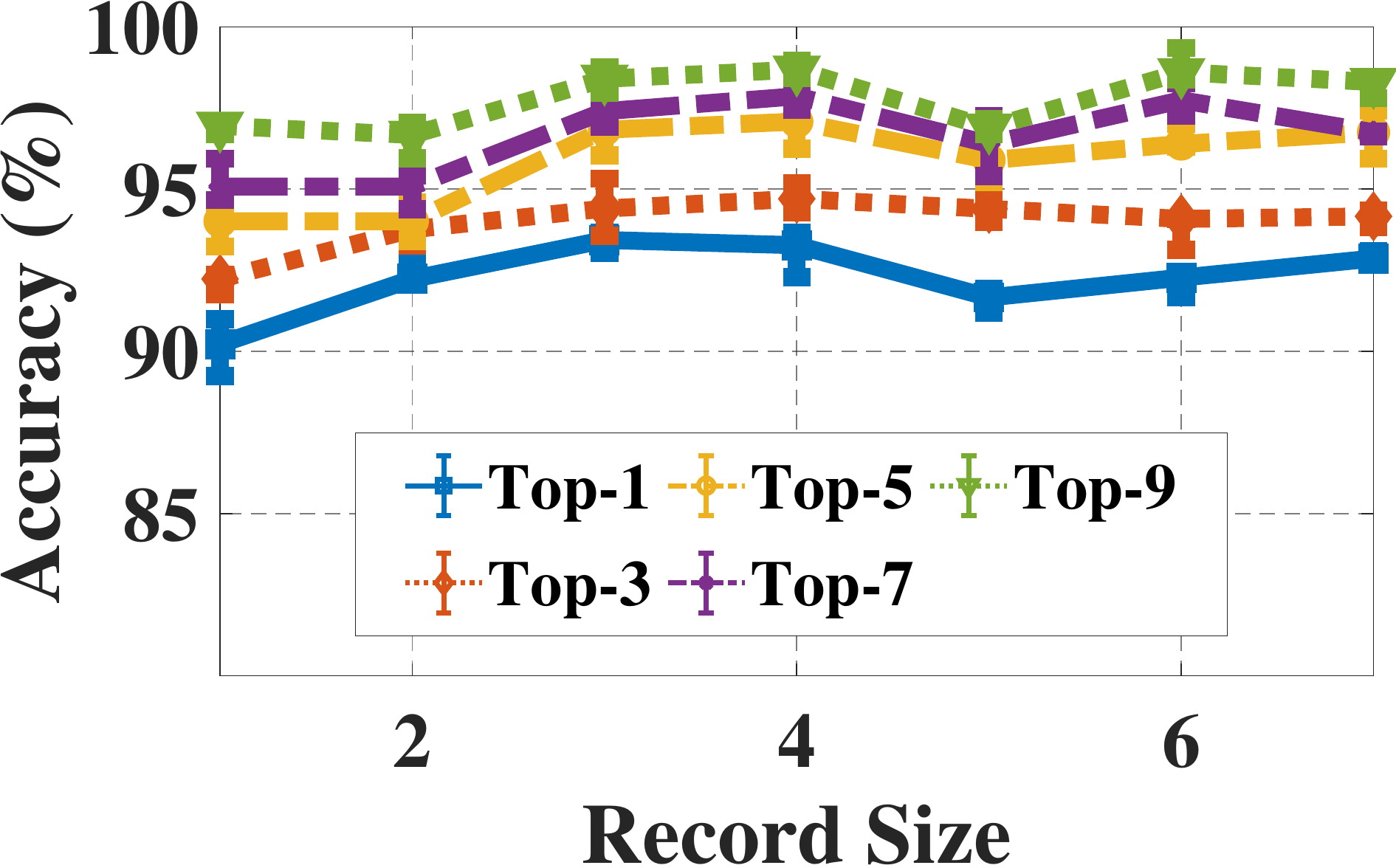}
        \vspace{-3mm}
        \caption{Impact of the RGB-D records per subject.}
        \label{fig:vision_sample}
        \vspace{-2mm}
    \end{minipage}
     \vspace{-1mm}
\end{figure}

We also evaluate the performance when the targets are walking away from the devices. Specifically, we additionally collect data of 15 candidates who asked to walk away from the devices during data collection. As Fig.\ref{fig:walkingaway} depicts, our design achieves $90.63\%$ top-1 accuracy, $93.75\%$ top-3,
and $96.88\%$ top-5 accuracy. The performance of our method outperforms the baseline approaches, indicating that our design is effective when the target is walking away from the devices.



\color{black}
\vspace{-2mm}
\subsubsection{Impact of the number of individual RGB-D records}
We change the number of records of a subject in the RGB-D database. This experiment mimics the situation in real life that cameras are able to capture one same subject multiple times and thereby forms a comprehensive candidate gallery. To this end, we randomly selected a different number of RGB-D records for each subject and analyze their accuracy from top-1 to top-9.
Fig.\ref{fig:vision_sample} shows the results for single-person settings. We found the availability of more RGB-D records for each subject can increase the chances of identifying the target person. This is reasonable because more records can mitigate the randomness in a person's walking style, e.g., the small changes of the walking speeds. 
\color{black}
However, even in the case where every subject is captured by camera only once, our method still achieves a robust result (90.22\% top-1 and 92.23\% top-3 accuracy). 
\color{black}


\begin{figure}[t]\vspace{-10mm}
    \centering
    \subfigure[Single subject.]
    {
        \begin{minipage}[t]{0.45\linewidth}
            \centering
            \includegraphics[width=0.85\linewidth]{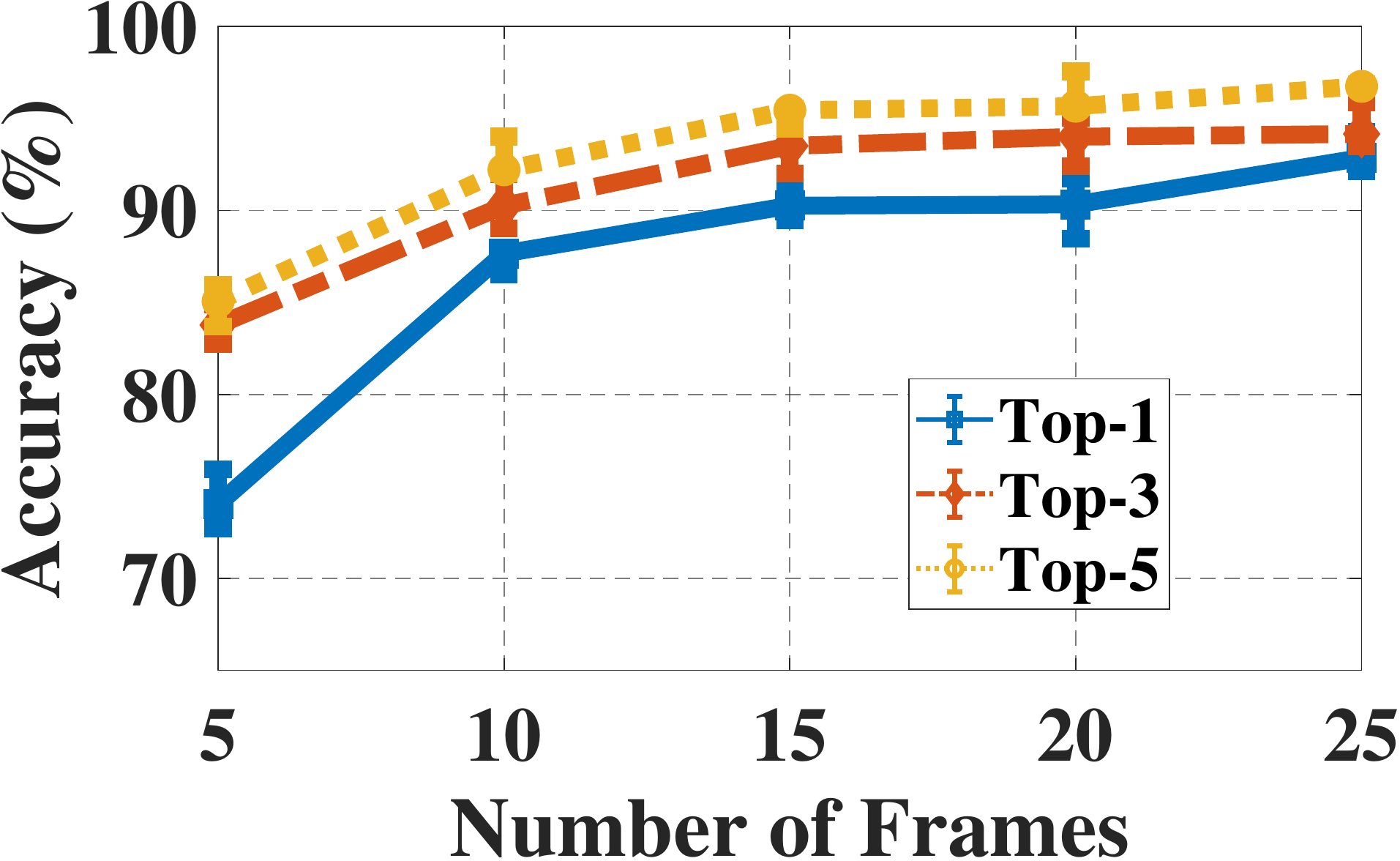}  
            %
            \label{}
        \end{minipage}
        \vspace{-5mm}
    }
    \subfigure[Multiple subjects.]
    {
        \begin{minipage}[t]{0.45\linewidth}
            \centering
            \includegraphics[width=0.85\linewidth]{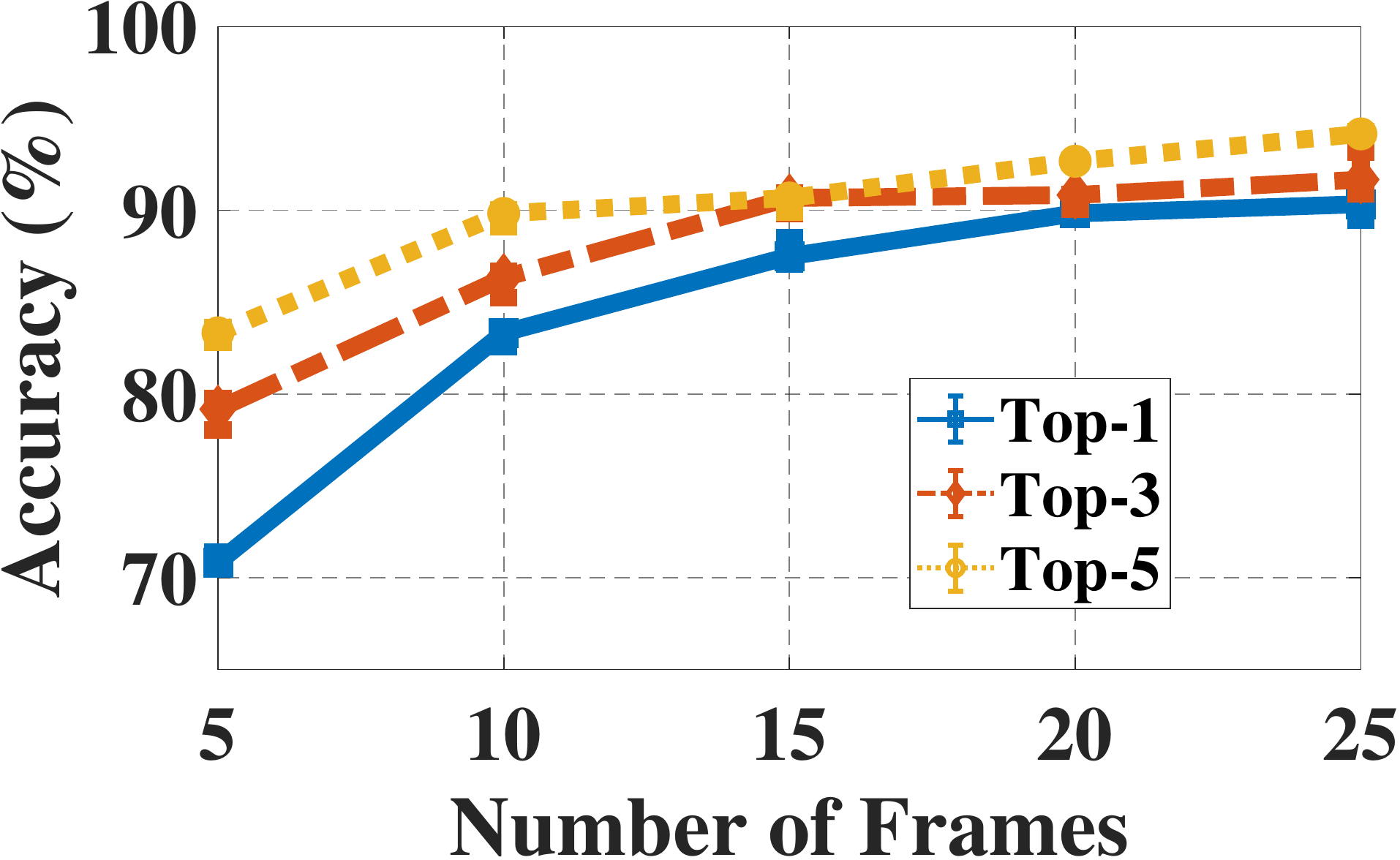}
            \label{}
            
        \end{minipage}
        \vspace{-5mm}
    }
    \vspace{-5mm}
    \caption{Impact of the number of gait frames.}
    \label{fig:diffCycles}
    \vspace{-5mm}
\end{figure}

\subsubsection{Impact of number of gait frames} \label{sec:frame_number}
Since the duration of a gait cycle is typically around 1 second, it is common that radar and RGB-D can provide the footage of the subject with more than one gait cycles. On the other hand, there are corner cases  (e.g., blockage) where only a fragment of gait cycle is captured.  Presumably, an increasing the number of gait frames can capture more identity-specific gait features and mitigate random errors. Thus, we repeat our evaluations by varying number of captured frames from 5 to 25 to simulate 0.5 to 2.5 cycles. Fig.\ref{fig:diffCycles} shows that when the number of gait frames is 25 (2.5 seconds),  the top-1 accuracy accuracy of single and multi-person ReID are  $92.86\%$ and $ 90.33\%$.  Since it is a very mild assumption in the real world to capture 2.5 second footage in each record, the results show that our model are accurate in most regular use cases. Even with half gait cycle, the model can still achieve $74.03\%$, showing its robust in  challenging situations.

\vspace{-1mm}
\subsubsection{Impact of the number of candidates}
\begin{figure*}[b]\vspace{-3.1mm}
  \includegraphics[width=0.7\linewidth]{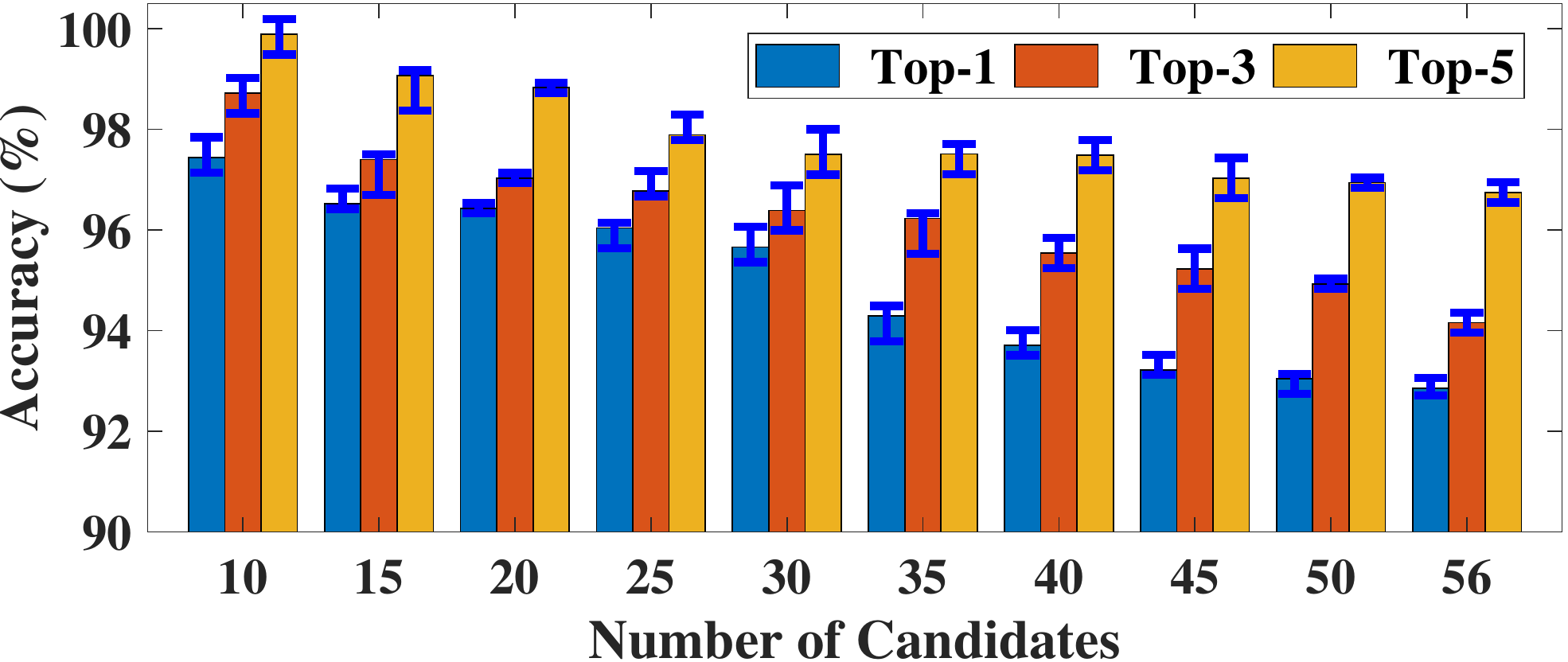}
  \vspace{-3.5mm}
  \caption{Impact of the number of candidates.}
  \label{fig:candidate}
  \vspace{-5.2mm}
\end{figure*}

The number of candidates in the RGB-D database impacts the performance. Presumably, the identification task becomes tougher with the growing number of candidates. Our overall performance is reported with 56 subjects while  in reality the number varies in the different scenarios. For example, in the domestic settings, the scale could be much smaller. Therefore, we further simulate the scenario where RGB-D captures different numbers of subjects in public areas. Fig.\ref{fig:candidate} shows that  the number of candidates impacts top-1 accuracy most. 
{\color{black}
For example, our method can achieve more than 97.4\% of the top-1 accuracy when handling 10 candidates. As the number of candidates increases to 56, the top-1 accuracy decreases to 92.86\%. In contrast, the top-2 and top-3 accuracy only decrease moderately on a large candidate set. Top-3 accuracy is kept above 94\% across all the settings.} We can thus provide the user of the system a confidence value of the identification result based on the number of candidates.

\color{black}

\color{black}
\vspace{-1mm}
\subsubsection{Impact of different mmWave radar scenes}
To evaluate the generalization performance of the model, we conduct experiments with radar data collected  in two different scenes as shown in Fig.\ref{fig:experiments} (a) and (b). In particular, radar data collected in scene 1 and 2 consists of 26 and 30 participants respectively while RGB-D gallery data is a complete set 56 participants.  The results show that the change of scenes has little effect on ReID accuracy. {\color{black} In scene 1, our method achieves accuracy rate top-1 93.16\% and top-5 96.06\%. In scene 2, our method achieves accuracy rate top-1 92.56\% and top-5 97.44\%.} The evaluation validates that the specular reflection signature of our design is resilient to environmental heterogeneity. The reason lies in that the mmWave sensor and our preprocessing algorithm is able to remove static reflection points from the different layout in different environments.

\color{black}

%% file: 10Related_work.tex
\UseRawInputEncoding
\vspace{-1.5mm}
\section{Related work}
\subsection{Single-modal Person Identification}
\vspace{-0.8mm}
Both vision-based and RF-based identification techniques have been extensively studied in the literature. In the computer vision, gait energy image (GEI) \cite{nambiar2019gait} is commonly utilized to extract a person's gait features (i.e., walking style) from RGB images. Depth information from RGB-D camera is also exploited to improve the identification accuracy \cite{wu2017robust,haque2016recurrent}. On the one hand, RF techniques (e.g., WiFi and radar) are proposed as privacy-preserving identification methods. WiFi CSI spectrum is analyzed in  \cite{7460727} as a gait feature for identifying a single user, while mmWave radars are adopted for multi-person identification. MU-ID \cite{yang2020mu} simultaneously recognizes up to 4 person by estimating their step lengths and duration from radar raw signal. mID \cite{zhao2019mid} trains a RNN model to concurrently track and identify 2 people via radar point clouds.

In contrast to these works, our design addresses an emerging problem of re-identifying people across vision and RF sensor, which gives rise to new challenges such as inter-modality discrepancy and misalignment.
\vspace{-2.5mm}
\subsection{Cross-modal Person Identification}
\vspace{-1mm}
{\color{black}
Cross-modal person identification has been studied by the computer vision community for years, though the majority of them focus on different types of cameras. These include the identification between depth and RGB images \cite{hafner2018cross, karianakis2018reinforced}, visible-to-infrared \cite{wu2017rgb,ye2018visible} and across different image resolution \cite{li2019recover}, which are only applicable at camera-allowed areas.
On the other side, despite its importance and ubiquity, the cross-modal ID between vision and RF signals receives little attention. 
XModal-ID \cite{korany2019xmodal} is the pioneering study of cross-modal ReID between the camera and RF.
However, XModal-ID is a CSI-based technique and struggles to simultaneously identify multiple people in the same FoV. 
In addition, XModal-ID requires a separate WiFi transmitter and receiver, which often introduces extra deployment cost and calibration complexity.
By investigating the emerging low-cost mmWave radars and RGB-D cameras, our proposed method leverages the scale and metric information provided by them and is able to achieve cross-modal ReID even in multi-person scenarios. The detailed comparison of cross-modal Re-ID system is shown in Table.\ref{tab:crossreid}. Our system unfolds the potential of cross vision-RF human sensing and seamless human identification across camera-allowed and camera-restricted areas.}

\renewcommand{\arraystretch}{2.2}\vspace{-3mm} 
\begin{table*}[t]
\small
\centering
\caption{Comparison of cross-modal Re-ID system.}
\vspace{-2mm}
\label{tab:crossreid}
\begin{tabular}{c|c c c c c c c}
\cline{1-8}
\hline 
\cline{1-8}

\textbf{System}
&{\makecell[c]{\textbf{Scenarios of}\\\textbf{the cross-modal}\\\textbf{combination}}}  
&\textbf{Accuracy}
&{\makecell[c]{\textbf{Field of view}\\\textbf{coverage of this}\\\textbf{combination}}}
&{\makecell[c]{\textbf{Multi-}\\\textbf{person}}}
&{\makecell[c]{\textbf{Visual}\\\textbf{privacy}\\\textbf{protection}}}
&{\makecell[c]{\textbf{Ease of}\\\textbf{ instal.}}}
&{\makecell[c]{\textbf{Power}\\\textbf{consum.}}}
\\
\hline
\cline{1-8}

{\makecell[c]{RGB -- IR \\\cite{wu2017rgb,ye2018visible}}}
&{\makecell[c]{Camera-allowed -- \\Camera-allowed}}
&High
&{\makecell[c]{Medium\\Wide}}
&Yes
&No
&Easy
&Low
 \\ 
\cline{1-8}

{\makecell[c]{RGB -- RGBD\\\cite{hafner2018cross, karianakis2018reinforced}}}
&{\makecell[c]{Camera-allowed -- \\Camera-allowed}}
&High
&{\makecell[c]{Medium\\Wide}}
&Yes
&No
&Easy
&Low
\\ 
\cline{1-8}

{\makecell[c]{RGB -- WiFi\\\cite{korany2019xmodal}}}
&{\makecell[c]{Camera-allowed -- \\Camera-restricted}}
&Medium
&Wide
&No
&Yes (WiFi)
&Hard
&Low

\\
\cline{1-8}
\cline{1-8}

{\makecell[c]{\textbf{RGBD -- Radar}\\\textbf{(Ours)}}}
&{\makecell[c]{\textbf{Camera-allowed --}\\\textbf{Camera-restricted}}}
&\textbf{High}
&\textbf{Wide}
&\textbf{Yes}
&\textbf{Yes (Radar)}
&\textbf{Easy}
&\textbf{Low}
\\  
\cline{1-8}
\hline
\cline{1-8}
\end{tabular}
\vspace{-4mm}
\end{table*}

\color{black}
\subsection{Multi-modal fusion}
\vspace{-1.5mm}
Our work is also related to multi-modal fusion between RF and vision sensor that has been investigated \cite{cao2018enabling, xu2019ivr, guo2018pedestrian} in the topics of location, detection and tracking, etc. However, our work is distinct from sensor fusion in two aspects. First, regarding application scenarios, these work focus on the problem where different sensors are installed or appear in the same site simultaneously whereas our work allows RF and camera to be installed in the disjoint areas to satisfy different camera restrictions. Second, in technical design, fusion works commonly differentiate and associate identities through the correlation in the location and trajectory  \cite{cao2018enabling}. In contrast, our work studies the problem of cross-modal human re-identification (ReID) between different places where deployed with heterogeneous sensors (i.g., cross-modal querying between different places).
Therefore, the challenges and technologies are different. First, we need to tackle the challenges of inter-modality discrepancy and robust similarity estimation  against practical factors (e.g., minor gait change and temporal misalignment).

\color{black}

%% file: 11Discussion.tex
\UseRawInputEncoding
\vspace{-3mm}
\section{Discussion AND FUTURE WORK}
\vspace{-1mm}
This work focuses on proof-of-principle cross vision-RF human re-identification by using low-cost mmWave radars and RGB-D cameras. There are limitations and future extensions.


\noindent \textbf{Number of people in the FoV.}
Constrained by the low-level HW/SW configurations of the mmWave radar used in this work (e.g., the maximum number of points per frame), we use a two-person experiment to demonstrate the effectiveness against multiple persons present in the sensors’ FoV. In future work, we will consider using more powerful commodity mmWave radars and evaluate our design on a greater number of people in the FoV.

\color{black}
\noindent \textbf{Blockage.}
During multi-person ReID, different persons and the mmWave device may be in a straight line, which would block each other in some frames. We observe there are  two categories of blockage scenarios: partial occlusion (some frames are not occluded in a sequence of gait frames) and complete occlusion (all the frames suffers from occlusion when subject is walking). As for the partial occlusion, we could use the gait frames that the subject is not occluded for ReID. These frames could be chosen using recent works of multi-subject localization and tracking with mmWave radar \cite{zhao2019mid}. Through multi-target trajectory tracking, the segments in which the subjects block each other can be detected, and these gait frames can be removed accordingly. The system performance under various number of gait frames is evaluated in Section \ref{sec:frame_number} - as shown in Fig.\ref{fig:diffCycles}, our method achieves a robust result (87\% top-1 and 90\% top-3 accuracy) when there are only 10 frames (1 second) available. The performance could drop when the number of frames is insufficient to capture the motion of one gait cycle.  
As for the complete occlusion which is fundamentally challenging to any kind of ReID systems due to insufficient sensory information, our proposed system cannot recognize the identities outside of its field of view. However, 
recent works \cite{zhou2019autonomous} propose multi-sensor architecture where multiple sensors are installed to form a complementary field of view. The multi-radar setup is outside the current scope of this proof-of-concept work, but in the future we plan to expand our current algorithm to multi-radar scenarios for the robust whole-scene ReID against blockage.
\color{black}

\noindent \textbf{Cross mmWave radar and RGB camera scenario.} We use low-cost RGB-D cameras as the visual modality, because they emerge as a promising sensor and can obtain the 3D information of walking humans. While, in real life, RGB cameras have been more widely deployed. Re-ID cross radars and RGB cameras will be further explored. We believe this is very feasible to extend as reconstructing a 3D mesh from a single or multiple RGB images has recently been an established topic \cite{pan2019deep}.


\color{black}

%% file: 12Conclusion.tex
\UseRawInputEncoding
\vspace{-3mm}
\section{Conclusion}
\vspace{-1mm}
This paper presents a novel system design that re-identifies a person across the data captured by mmWave radar and RGB-D cameras. 
{\color{black}
To address the fundamental data discrepancy across heterogeneous sensors, we proposed a novel signature synthesis method based on the observed specular reflection model.
Our system also features an effective cross-modal deep metric learning method to handle the interference factors caused by unsynchronized data across radars and cameras.}
We believe our system unfolds the potential of cross vision-RF human sensing and envision it to serve as a key solution to seamless human identification across \textcolor{black}{camera-allowed and camera-restricted} areas.  
 